\pdfoutput=1

\documentclass[11pt]{article}
\usepackage{rotating}
\usepackage[final]{acl}
\usepackage{float}
\usepackage{times}
\usepackage{latexsym}
\usepackage{booktabs}
\usepackage{graphicx}
\usepackage{multirow}
\usepackage{adjustbox}
\usepackage{stfloats}
\usepackage{enumitem}
\usepackage{xurl}
\usepackage[T1]{fontenc}
\usepackage{longtable}

\usepackage{amsmath}

\usepackage[utf8]{inputenc}
\usepackage{cleveref}
\crefformat{section}{\S#2#1#3}

\usepackage{microtype}

\usepackage{inconsolata}

\usepackage{graphicx}

\usepackage{multirow}
\usepackage{tcolorbox}
\usepackage{enumitem}

\usepackage{tabularx}
\usepackage{array}
\usepackage{ragged2e}

\usepackage{listings}
\usepackage{tcolorbox}
\tcbuselibrary{listings, minted}
\usepackage{minted}
\usepackage{makecell}

\usepackage{placeins}
\usepackage{hyperref}

\newcolumntype{Y}{>{\RaggedRight\arraybackslash}X}
\newcommand{\tblsec}[1]{%
  \addlinespace
  \multicolumn{3}{@{}l}{\textbf{#1}}\\
}

\title{Annotation-Efficient Vision-Language Model Adaptation to the Polish Language Using the LLaVA Framework}

\author{
 \textbf{Grzegorz Statkiewicz},\quad
 \textbf{Alicja Dobrzeniecka},\quad
 \textbf{Karolina Seweryn},\\
 \textbf{Aleksandra Krasnodębska},\quad
 \textbf{Karolina Piosek},\quad
 \textbf{Katarzyna Bogusz}, \\
 \textbf{Sebastian Cygert},\quad
 \textbf{Wojciech Kusa}
 \vspace{0.3em}
\\
 NASK National Research Institute, Warsaw, Poland
\\
 \small{
   \textbf{Correspondence:} \texttt{\{firstname.lastname\}@nask.pl} 
 }
}

\begin{document}
\maketitle
\begin{abstract}
Most vision-language models (VLMs) are trained on English-centric data, limiting their performance in other languages and cultural contexts. 
This restricts their usability for non-English-speaking users and hinders the development of multimodal systems that reflect diverse linguistic and cultural realities.
In this work, we reproduce and adapt the LLaVA-Next methodology to create a set of Polish VLMs. We rely on a fully automated pipeline for translating and filtering existing multimodal datasets, and complement this with synthetic Polish data for OCR and culturally specific tasks. Despite relying almost entirely on automatic translation and minimal manual intervention to the training data, our approach yields strong results: we observe a +9.5\% improvement over LLaVA‑1.6‑Vicuna‑13B on a Polish-adapted MMBench, along with higher-quality captions in generative evaluations, as measured by human annotators in terms of linguistic correctness. These findings highlight that large-scale automated translation, combined with lightweight filtering, can effectively bootstrap high-quality multimodal models for low-resource languages. 
Some challenges remain, particularly in cultural coverage and evaluation. To facilitate further research, we make our models and evaluation dataset publicly available.
\end{abstract}


\section{Introduction}
Recent advances in artificial intelligence have led to remarkable progress in multimodal large language models (LLM), especially vision-language models (VLMs), which integrate vision and language understanding to perform tasks such as visual question answering, image captioning, and reasoning~\cite{achiam2023gpt,liu2023llava,liu2024llavanext}. These models leverage massive datasets and sophisticated architectures to achieve state-of-the-art performance across a wide range of benchmarks. 
However, the current VLM landscape remains predominantly English-centric, primarily due to the composition of standard training datasets, which limits effectiveness in other languages and cultural contexts~\cite{tong2024cambrian1, laurençon2024matters, wiedmann2025finevisionopendataneed}. 

To address this challenge, we explore whether large-scale automated translation can serve as a practical alternative for developing multimodal models in low- and mid-resource languages. Specifically, we focus on Polish as a case study and investigate how far we can go using automatically translated data with minimal manual intervention in the training data.
We choose Polish due to the availability of recent competitive Polish LLMs~\cite{kocon2025pllum,ociepa2025bielik} and its rich morphological complexity, which poses challenges for both text and multimodal understanding.

\begin{figure}[t]
    \centering
    \includegraphics[width=\linewidth]{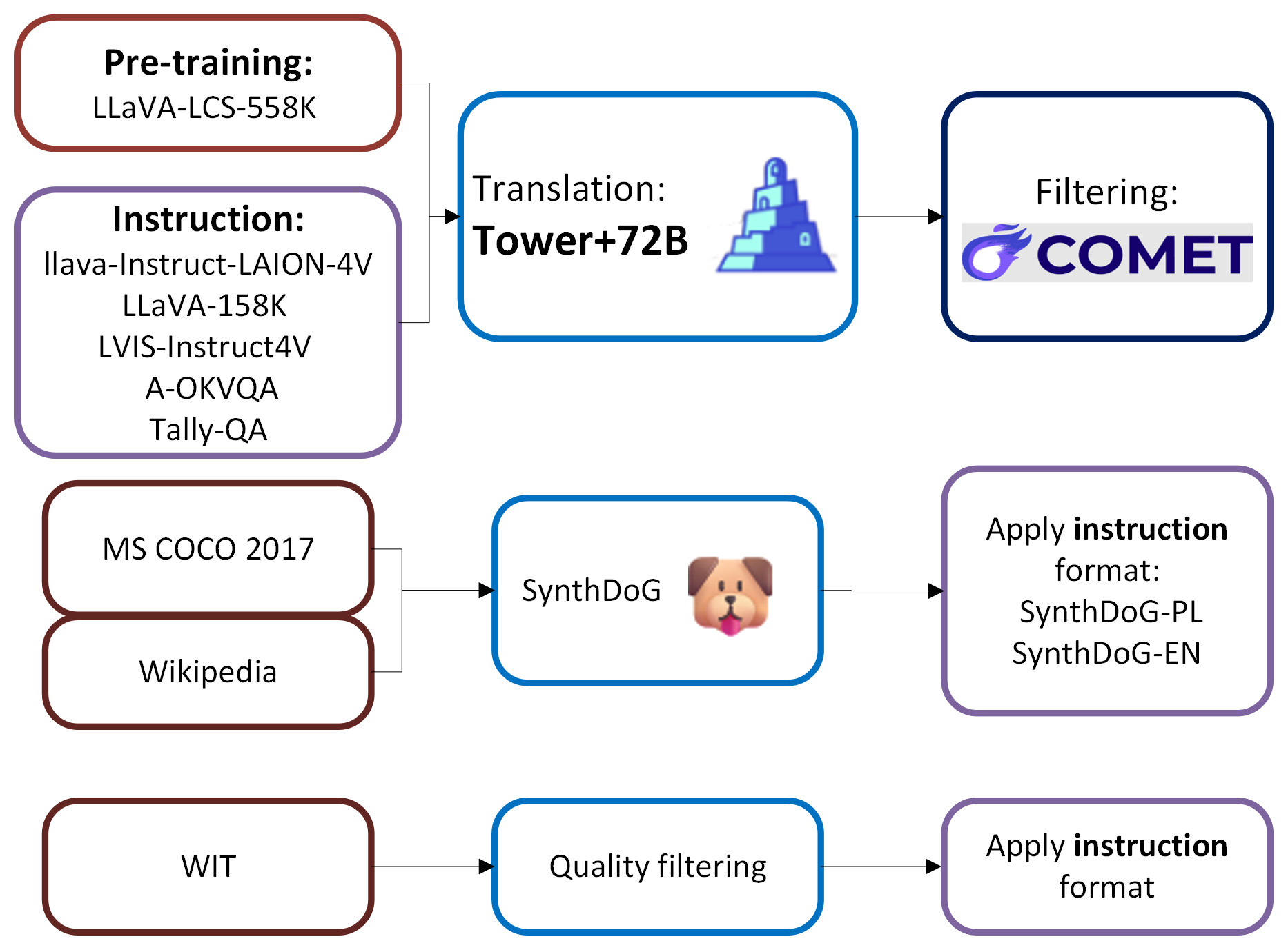}
    \caption{Our custom dataset construction process.}
    \label{fig:dataset_construction}
\end{figure}

Our pipeline employs Tower+ 72B~\cite{rei2025towerplus}, a state-of-the-art multilingual model, to translate popular multimodal datasets for both pretraining and instruction tuning, which include general visual question answering (VQA), synthetic optical character recognition (OCR), and counting tasks (see Figure~\ref{fig:dataset_construction} for overview). For rigorous evaluation, we also translate the MMBench dataset~\cite{Liu2023MMBench} and subject it to comprehensive human revision, ensuring high-quality benchmarks. 

The model we propose builds on the LLaVA-NeXT architecture~\cite{liu2024llavanext}, which aligns a pretrained visual encoder with an LLM via a lightweight two-layer MLP projector. For the language backbone, we use two variants of the \textsc{PLLuM-12B} model~\cite{kocon2025pllum} and the \textsc{Bielik-11B} model, all of which are Polish-native, instruction-tuned LLMs. For the vision tower, we replace the CLIP-like encoder~\cite{Radford2021CLIP} commonly used in LLaVA with SigLIP2~\cite{tschannen2025siglip2multilingualvisionlanguage}, chosen for its strong multilingual image-text alignment and robust text-localization signal.

Evaluations on the Polish-adapted MMBench show considerable improvements over baseline models of LLaVA family, in both Polish and English versions of the dataset. 
Additionally, we conduct LLM- and VLM-as-a-judge evaluations, along with manual assessment, and demonstrate that our model matches or surpasses state-of-the-art open-access models (PaliGemma2-10B~\cite{beyer2024paligemma}, Pixtral-12B~\cite{agrawal2024pixtral}, and Qwen2.5-VL-7B~\cite{yang2024qwen2.5}) in generating linguistically correct Polish captions.
Overall, the contributions of this paper are as follows:
\begin{itemize}[topsep=2pt, partopsep=0pt, itemsep=0pt, leftmargin=20pt]
    \item We present a fully automated pipeline for preparing multimodal datasets for low-resource languages, including translation, filtering, and quality estimation, complemented by synthetic data for tasks that are difficult to translate (e.g., OCR).
    \item  We introduce a family of Polish vision-language models (LLavA-PLLuM and LLaVA-Bielik) trained using the above dataset with a LLaVA-Next architecture and Polish-native LLM backbones (11B–12B parameter range).
    \item We conduct a comprehensive adaptation of the MMBench-dev dataset, alongside linguistic re-annotation of the corpora, identifying issues in nearly 4\% of the samples.
   \item We release our models and Polish version of MMBench-dev to support future research in multilingual multimodal LLMs.\footnote{\href{https://huggingface.co/collections/NASK-PIB/llava-pllum}{https://huggingface.co/collections/NASK-PIB/LLaVA-PLLuM}}
\end{itemize}

We start by describing the related work (\S\ref{sec:related}). 
Then, \S\ref{sec:model} introduces the 
model architecture, \S\ref{sec:data} presents datasets used for training and evaluation, and \S\ref{sec:experiment} details the experimental setup. 
Finally, \S\ref{sec:results} reports the results and discusses the findings.

\begin{figure*}[!ht]
    \centering
    \includegraphics[width=0.94\textwidth]{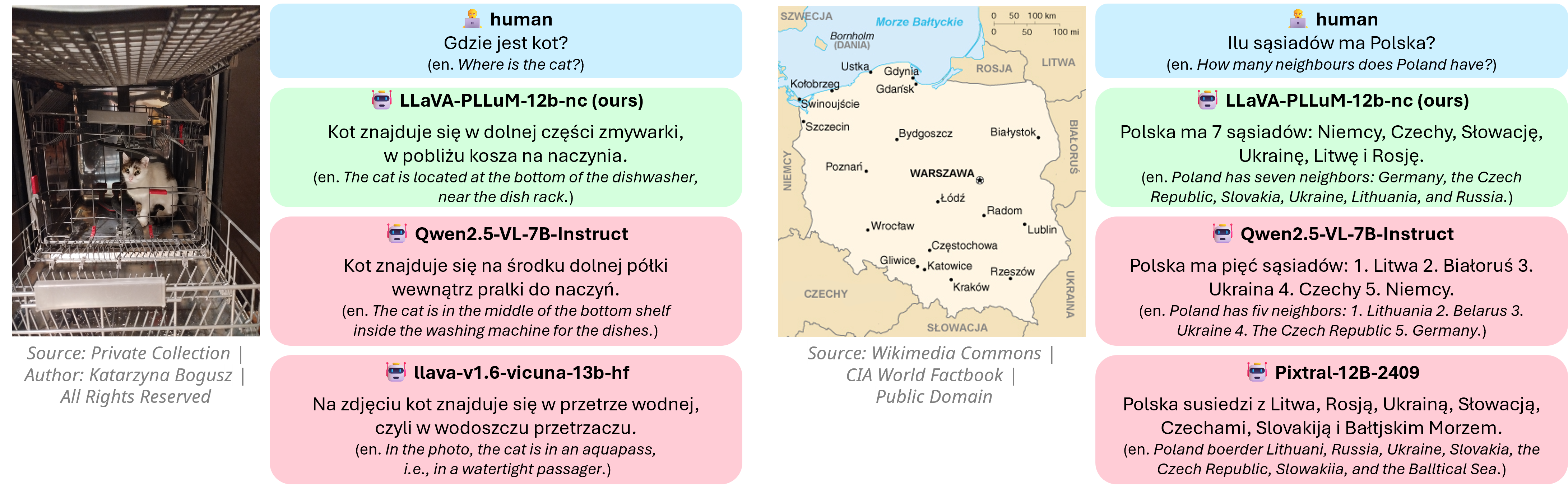}
    \caption{Comparative analysis of sample VLM predictions on two example images from our internal evaluation dataset. For each image, the human-provided prompt is shown, followed by our model and other baseline models' predictions. All predictions are presented in Appendix~\ref{app:internal}.}
    \label{fig:qualitive_exaples}
\end{figure*}

\section{Related Work} \label{sec:related}

In this section, we review prior work on vision-language models and their adaptation to multilingual and language-specific settings, providing context for our approach.

\subsection{Vision-language models}

The field of NLP has evolved with the introduction of large language models (LLMs) \cite{ouyang2022training,achiam2023gpt,touvron2023llama}. 
In parallel, recent research has sought to extend these models beyond text by incorporating additional modalities, giving rise to vision-language models (VLMs) \cite{zhang2024vision}. Early efforts in this direction focused on learning aligned representations between visual and linguistic inputs \cite{alayrac2022flamingo}. Notable examples include CLIP \cite{Radford2021CLIP} and ALBEF \cite{li2021align}, which employ contrastive learning objectives to bridge image and text representations. Building on this foundation, subsequent models such as BLIP \cite{li2022blip}, BLIP-2 \cite{li2023blip2} and SigLIP2~\cite{tschannen2025siglip2multilingualvisionlanguage} introduced more advanced pretraining and architectural designs to enhance cross-modal reasoning and generation capabilities.

Several recent studies have explored adapting the LLaVA architecture to support specific target languages and multilingual settings.
\citet{shin-etal-2024-x} focus on Korean-English use cases and propose X-LLaVA, an extension of the LLaVA-1.5 framework. Their approach incorporates three key components: (1) vocabulary expansion for the target language by adding Korean tokens to the LLaMA-2 model; (2) cross-lingual pretraining to connect knowledge across multiple languages \cite{10.5555/3454287.3454921}; and (3) multilingual visual instruction tuning, which combines machine-translated versions of the LLaVA instruction dataset with their newly generated MVIF dataset.

\citet{Musacchio2024LlavaNdino} propose LLaVA-ndino, which adapts the LLaVA framework for Italian using machine-translated datasets. Their training pipeline divides visual instruction tuning into two stages: the first stage enhances performance on vision-language tasks by incorporating The Cauldron dataset \cite{laurençon2024matters}, while the second stage focuses on generating longer and more coherent responses through additional training on the LLaVA Conversation dataset.

Similarly, \citet{alam2024mayainstructionfinetunedmultilingual} build a multilingual vision-language model based on the LLaVA architecture by translating LLaVA datasets into eight languages. In \citet{Li_Yang_Jiang_Zhu_Sun_2025}, the authors address low-resource language scenarios and propose LRM-LLaVA, which integrates a cross-modal regularizer alongside translated datasets to improve performance in low-resource settings.

 In contrast to prior work, our approach focuses on a single morphologically rich target language and relies almost entirely on large-scale automated translation paired with Polish-native LLM backbones, without vocabulary modification or cross-lingual pretraining, and with human intervention limited to evaluation benchmark curation.

\subsection{Visual instruction following datasets}

The development of vision-language models has been strongly driven by the availability of large-scale, high-quality multimodal datasets. Early datasets such as MS-COCO \cite{10.1007/978-3-319-10602-1_48} provided image-text pairs and object-level annotations, forming the basis for image captioning and visual grounding tasks. More recent efforts have focused on instruction-style datasets, which better support conversational and reasoning-oriented VLMs. Notable examples include LLaVA-Instruct \cite{liu2023llava}, constructed by augmenting image-text pairs with GPT-generated instructions, The Cauldron \cite{laurençon2024matters}, which aggregates diverse vision-language datasets into a unified training resource, and WIT (Wikipedia-based Image-Text) \cite{10.1145/3404835.3463257}, a large-scale multilingual dataset designed to support cross-lingual and cross-modal learning.

To evaluate the performance of vision-language models, several standardized benchmarks have been proposed. MM-Bench \cite{Liu2023MMBench} is a multiple-choice benchmark designed to assess multimodal perception, reasoning, and knowledge across a wide range of vision-language tasks. In contrast, XM3600 \cite{ThapliyalCrossmodal2022} focuses on multilingual image-text understanding, providing image-caption pairs and retrieval-style evaluations across diverse languages and cultural contexts.

\subsection{Polish large language models}

From a linguistic perspective, Polish poses several challenges for large language models. It is a highly inflected language with rich morphology, including seven grammatical cases, complex agreement patterns, and relatively free word order. These properties increase surface-form sparsity and complicate both generation and understanding, particularly for tasks requiring precise grammatical agreement or fine-grained semantic distinctions. As a result, Polish serves as a meaningful testbed for studying multilingual and low-resource adaptations of large language models.

The Polish NLP ecosystem has recently seen the development of several LLMs specifically designed for the language. Prominent examples include \textit{PLLuM}~\cite{kocon2025pllum} and \textit{Bielik}~\cite{ociepa2025bielik}, which are available in both base and instruction-tuned variants and are optimized to handle the syntactic, morphological, and semantic properties of Polish.
Recent efforts have further emphasized instruction tuning using large-scale, Polish-specific supervision.
Notable resources include \textit{PLLuM-Align}~\cite{seweryn2025pllum} and \textit{PLLuMIC}~\cite{pkezik2025pllum}, which provide high-quality instruction and conversation-style datasets constructed from a mixture of translated, synthetic, and manually curated data.

Despite this progress in text-only modeling, to the best of our knowledge there are no existing datasets or models that directly support Polish in vision-language research. 
Our work addresses this gap by introducing Polish-adapted VLMs and benchmarks, with human annotation restricted to evaluation to ensure scalability and reproducibility.

\section{Model} \label{sec:model}


We build on the LLaVA-NeXT architecture~\cite{liu2024llavanext}, which aligns a pretrained visual encoder with an LLM via a lightweight two-layer MLP projector. This design preserves the LLM's strong language prior while enabling efficient multimodal grounding. Compared to the original LLaVA~\cite{liu2023llava}, LLaVA-NeXT supports higher input resolutions and dynamic tiling, features that have been observed to improve fine-grained perception and OCR performance.

As the language backbone, we use three leading Polish-native, instruction-tuned LLMs within the 11--12B parameter size range to evaluate their effectiveness in multimodal settings:
\begin{itemize}
    \item \textsc{PLLuM-12b-nc-instruct-250715}\footnote{\url{https://huggingface.co/CYFRAGOVPL/pllum-12b-nc-instruct-250715}}
    \item \textsc{Bielik-11B-v2.6-Instruct}\footnote{\url{https://huggingface.co/speakleash/Bielik-11B-v2.6-Instruct}}
    \item \textsc{PLLuM-12B-nc-instruct}\footnote{\url{https://huggingface.co/CYFRAGOVPL/PLLuM-12B-nc-instruct}}
\end{itemize}
For the vision tower, we replace the CLIP-like encoder commonly used in LLaVA variants with \textsc{SigLIP2 So400m/14, 384px}~\cite{tschannen2025siglip2multilingualvisionlanguage}, selected for its stronger multilingual image-text alignment and more robust text-localization signal.

We adopt a two-stage training strategy. In the first stage, we update only the projector to align visual features with the LLM's embedding space using image-caption pairs. The second stage involves jointly fine-tuning the projector and vision encoder, while the language model is updated via LoRA \cite{hu2022lora} on a diverse set of multimodal instructions. We prioritize this parameter-efficient strategy over full fine-tuning to optimize computational resource usage while preserving the backbone's pre-trained linguistic competencies.
We employ high-rank adapters ($r=128, \alpha=256$) to ensure sufficient capacity for cross-modal alignment. 
The details of the specific configuration for both stages are provided in Appendix~\ref{app:training-details}.

\section{Data} \label{sec:data}

Building effective non-English VLMs is often hindered by the lack of high-quality, language-specific multimodal instruction data.
To address this for Polish, we implement a scalable pipeline combining large-scale automated translation, metric-based filtering, and synthetic data generation. 
In this section, we detail the construction of our training mixtures for general visual reasoning, OCR, and cultural knowledge, and describe our careful linguistic and content-based refinement of evaluation benchmarks to ensure reliable assessment.

\begin{table}[t]
\centering
\small
\renewcommand{\arraystretch}{1.1}
\begin{tabularx}{\linewidth}{@{}l r Y@{}}
\toprule
\textbf{Category} & \textbf{\# Samples} & \textbf{Sources} \\
\midrule
General   & 906K    & Allava-Instruct-LAION-4V; LLaVA-158K; Q-Instruct; LVIS-Instruct4V; A-OKVQA \\
OCR       & 600K    & SynthDoG-PL; SynthDoG-EN \\
Knowledge & 390K    & WIT \\
Counting  & 104K    & TallyQA \\
\midrule
\textbf{Total} & \textbf{2.0M} & \\
\bottomrule
\end{tabularx}
\caption{Instruction data mixture by category. Specified counts determine the number of conversations.}
\label{tab:inst-composition}
\end{table}

 \subsection{VLM-training datasets}

To meet the Polish-first design goals, we construct distinct datasets corresponding to the two-stage training process. 
Specifically, we prepare a captioning corpus for pre-training and a comprehensive visual instruction tuning mixture to develop diverse multimodal capabilities. 

\subsubsection{Polish language adaptation} \label{data-adaptation}

We adapt English multimodal conversations to Polish using a three-stage pipeline: translation, quality estimation, and filtering. Each sample is a multi-turn dialogue decomposed into interleaved question-answer pairs. Every sample is translated with \textsc{Tower+ 72B}~\cite{rei2025towerplus}. Translation quality is assessed with reference-less COMET metric~\cite{rei-etal-2020-unbabels}. If either side of a QA pair falls below a fixed threshold, the pair is removed and dialogues with no remaining pairs are discarded. Based on preliminary manual inspection, the threshold can vary from 0.4 up to 0.8 depending on the source dataset.

\subsubsection{Pre-training data} \label{pretrain-data}

For the pre-training (feature-space alignment phase) we use the \emph{LLaVA-LCS-558K} corpus  with $595$K multi-turn samples, which provides broad image-text coverage with simple conversational templates~\cite{liu2023llava}. 
We apply the adaptation pipeline described in Section~\ref{data-adaptation} to obtain a Polish-majority mixture, ensuring that the visual tokenization and projector learn to interface with Polish prompts early in training.

\subsubsection{Instruction data} \label{instruct-data}

Below we describe the instruction mixture which is constructed to cover a wide range of capabilities under four categories.
To preserve English language capabilities, the resulting instruction set maintains an 85:15 balance between Polish-adapted and original English samples.
The composition of the final instruction dataset is summarized in Table~\ref{tab:inst-composition}. 

\paragraph{General} \label{instruct-general}

We aggregate general-purpose VLM supervision from \emph{Allava-Instruct-LAION-4V}~\cite{chen2024allavaharnessinggpt4vsynthesizeddata}, \emph{LLaVA-158K}~\cite{liu2023llava}, \emph{Q-Instruct}~\cite{wu2023qinstruct}, \emph{LVIS-Instruct4V}~\cite{wang2023instruct4v}, and \emph{A-OKVQA}~\cite{schwenk2022aokvqabenchmarkvisualquestion}. Each dataset is processed as in Section~\ref{data-adaptation}.

\paragraph{OCR} \label{instruct-ocr}

To emphasize reading ability, we synthesize OCR-centric conversations following the \textsc{SynthDoG} procedure~\cite{kim2022donut}. Text snippets are sampled from Polish and English Wikipedia, typeset with randomized fonts, sizes, and placements, and composited onto natural-image backgrounds drawn from \textsc{MS~COCO~2017}~\cite{10.1007/978-3-319-10602-1_48}. For each image we instantiate an instruction from a hand-crafted template set (e.g., ``Przepisz widoczny tekst / Read the text shown''), and set the answer to the rendered string. This produces two datasets---\emph{SynthDoG-PL} and \emph{SynthDoG-EN}---covering receipts, signs, labels, and document-like layouts with diacritics and mixed-case patterns.

\paragraph{Knowledge} \label{instruct-knowledge}

To inject factual and cultural grounding, we derive image-question pairs from the Wikipedia-based Image Text (WIT) dataset~\cite{srinivasan2021wit}. We retain only Polish samples with human-written captions, and convert each into a single-turn conversation by sampling an instruction template (e.g., ``Opisz obraz / Describe the image'') and using the caption as the target response. 

\paragraph{Counting} \label{instruct-counting}

Finally, we include instances from \emph{TallyQA}~\cite{acharya2019tallyqa} to explicitly train numeracy and set-size reasoning in natural scenes. Prompts are translated and filtered as in Section~\ref{data-adaptation}.

\subsection{MMBench adaptation} \label{sec:mmbench-adaptation}

To create the first Polish-language instruction vision evaluation dataset, we start with the English MMBench dev set. 
MMBench is a comprehensive multi-choice benchmark designed to systematically evaluate diverse capabilities such as fine-grained perception and logical reasoning, making it a robust standard for assessing general-purpose vision-language models.
We first machine-translate it into Polish using the TOWER+ 72B model. 
Subsequently, native Polish professional linguists, employed full-time by our organization, perform a thorough review of the translated output, making both linguistic and content-related corrections.

During this process, two main types of issues were identified: (1) linguistic or content inaccuracies, and (2) questions requiring foreign cultural or linguistic knowledge. Overall, 3.56\% of the questions contained inaccuracies, while additional 3.02\% were rooted in foreign contexts, out of a total of 1,292 questions. 
To ensure a reliable evaluation, problematic questions were corrected when possible during the adaptation.
Detailed categorization of these issues, the percentage of affected questions, and the handling of questions requiring foreign cultural knowledge, is presented in Appendix~\ref{app:mmbench}.

\section{Experiment setup} \label{sec:experiment}

This section describes the experimental setup, including model training, baseline selection, and evaluation protocols.

\subsection{Model training}

We train three described models using datasets and procedure described in Sections~\ref{sec:model}--~\ref{sec:data}.
We perform model training using high-performance computing clusters equipped with NVIDIA A100 (40GB) and NVIDIA GH200 (96GB) accelerators. Specifically, the pre-training phase (Stage 1) is executed on A100 GPUs, requiring approximately 336 GPU-hours per model. The visual instruction tuning phase (Stage 2) uses GH200 nodes and consumes approximately 1,344 GPU-hours per model.

\subsection{Baselines}

We evaluate our method against five open-weight VLMs of comparable size, selected to provide two complementary perspectives.
To establish architectural baselines, we employ \textsc{LLaVA-1.6-Mistral-7B}\footnote{\url{https://huggingface.co/llava-hf/llava-v1.6-mistral-7b-hf}}~\cite{liu2023improved} and \textsc{LLaVA-1.6-Vicuna-13B}\footnote{\url{https://huggingface.co/liuhaotian/llava-v1.6-vicuna-13b}}. 
Additionally, we compare against \textsc{Qwen2.5-VL-7B}\footnote{\url{https://huggingface.co/Qwen/Qwen2.5-VL-7B-Instruct}}~\cite{Qwen2VL}, \textsc{PaliGemma2-10B}\footnote{\url{https://huggingface.co/google/paligemma2-10b-mix-224}}~\cite{paligemma2_2024}, and \textsc{Pixtral-12B}\footnote{\url{https://huggingface.co/mistralai/Pixtral-12B-2409}}, which represent state-of-the-art open-access models, allowing us to assess our models' competitiveness against leading multilingual systems.

\subsection{Evaluation procedure}

We evaluate our models both on a representative multimodal benchmark, as well as image captioning quality. The following subsections describe the datasets and corresponding evaluation protocols.

\subsubsection{MMBench} 
We evaluate our models on the \textit{MMBench V1.1} benchmark~\cite{Liu2023MMBench}, using the \emph{dev} split to ensure local reproducibility. To assess both cross-lingual transfer capabilities and native alignment, we conduct evaluations on two linguistic variants: the original English set and the Polish-adapted version described in Section~\ref{sec:mmbench-adaptation}. 

We use a direct response generation strategy followed by rule-based extraction. Specifically, we prompt the model to output the answer choice directly and use regular expression matching to map the generated text to one of the predefined options (A, B, C, or D). We report an accuracy as our primary metric.

\subsubsection{XM3600} 

XM3600~\citep{ThapliyalCrossmodal2022} is a dataset for image captioning, containing diverse real-world photographs showing everyday objects, people, activities, and scenes in varied indoor and outdoor environments, designed to evaluate visual understanding.
Since the task is generative, simple accuracy metrics are insufficient. We evaluate our models on XM3600 using three complementary approaches:

\begin{enumerate}[left=5pt]
    \item \textbf{Open-source LLM and VLM judges} using Llama-3.3-70B-Instruct and LLaVA-OneVision-Qwen2-7B-SI-HF. This evaluation was conducted on the full XM3600 dataset (3600 images), providing large-scale automatic assessment of generative caption quality. Judges were prompted to choose which model performed better, with no possibility of a tie.
    
    \item \textbf{Closed-source VLM judge} using Claude Sonnet 4.5. For this evaluation, we selected a representative sample of 500 images. Descriptions generated by our models and the two best performing baseline models from our experiments (Pixtral-12B-2409 and Qwen2.5-VL-7B-Instruct) were compared in a pairwise setup, allowing for three possible outcomes: a win for model A, a win for model B, or a tie.
    
    \item \textbf{Manual evaluation} by native Polish professional linguists. Due to task complexity, a subset of 400 image-caption pairs was used (details in  Appendix~\ref{app:manaul_numer_of_samples}). Each item was evaluated by a single annotator; before starting, all annotators jointly scored a set of 10 items to calibrate the annotation interface and ensure consistent judgments.
    The two best-performing models from our experiments were compared against the two best-performing baselines. Similarly to Claude evaluation, annotators judged which description was better or whether the result constituted a tie.
\end{enumerate}

Judges and human annotators determined which model performed better or whether the results constituted a tie according to two criteria: (1) \textbf{linguistic correctness}, considering the absence of grammatical, orthographic, punctuation, and syntactic errors, as well as a natural and fluent style in Polish with correct phraseology and no calques from English or incorrect word formation; and (2) \textbf{content description quality}, evaluated independently of linguistic form, focusing solely on faithfulness to the image content (absence of hallucinations), correct identification of key scene elements, and accurate representation of the environment, objects, actions, and relations between them.
Evaluation prompts are detailed in Appendix~\ref{app:eval_prompts}. 

\section{Results and Discussion} \label{sec:results}

\begin{table}[t]
    \centering
    \small
    \resizebox{\columnwidth}{!}{
        \begin{tabular}{lrr}
            \toprule
             &  \multicolumn{2}{c}{\textbf{MMBench V1.1 DEV}} \\
            \midrule
            \textbf{Model} & \textbf{PL} \hspace{-5pt} & \textbf{EN} \\
            \midrule
            LLaVA-1.6-Mistral-7B & 68.18 & 76.54 \\
            LLaVA-1.6-Vicuna-13B & 69.80 & 74.39 \\ \midrule
            LLaVA-PLLuM-12b-nc-250715 (Ours) & 76.73 & 75.23 \\ 
            LLaVA-Bielik-11b-v2.6 (Ours) & 78.24 & 77.75 \\
            LLaVA-PLLuM-12b-nc (Ours) & \textbf{79.35} & \textbf{78.43} \\
            \midrule
            \midrule
            Qwen2.5-VL-7B & 75.56 & 80.62 \\
            PaliGemma2-10B & 78.39 & 80.46 \\
            Pixtral-12B & \underline{82.06} & \underline{84.31} \\
            \bottomrule
        \end{tabular}
    }
    \caption{Comparison of model accuracy (\%) on MMBench V11 DEV. \textbf{Bold} denotes best result from LLaVA architecture-based models, \underline{underline} is best overall.
    }
    \label{tab:mmbehcnresults}
\end{table}

\begin{table}[t]
   \centering
    \small
    \resizebox{\columnwidth}{!}{
        \begin{tabular}{lccc}
            \toprule

  & \rotatebox{60}{\makecell{LLaVA-PLLuM\\12B-nc-250715}} 
 & \rotatebox{60}{\makecell{LLaVA-PLLuM\\12B-nc}} 
 & \rotatebox{60}{\makecell{LLaVA-Bielik\\11B-v2.6}} \\

\midrule
LLaVA-1.6-Mistral-7B & \textbf{84.91} & \textbf{85.81} & \textbf{82.35} \\
LLaVA-1.6-Vicuna-13B & \textbf{63.64} & \textbf{66.71} & \textbf{60.32} \\
PaliGemma2-10B & \textbf{77.47} & \textbf{77.53} & \textbf{74.1} \\
Pixtral-12B & 43.38 & 48.33 & 40.31 \\
Qwen2.5-VL-7B & 42.69 & 43.15 & 34.76 \\
\bottomrule
\end{tabular}
}
\caption{Preference rate (\%) of our models over baseline models judged by LLM (Llama-3.3-70B-Instruct) on XM3600 dataset for \textit{linguistic correctness} of descriptions.}
 \label{tab:xm3600_results_llm}
\end{table}

\begin{table}[t]
   \centering
    \small
    \resizebox{\columnwidth}{!}{
        \begin{tabular}{lccc}
            \toprule

  & \rotatebox{60}{\makecell{LLaVA-PLLuM\\12B-nc-250715}} 
 & \rotatebox{60}{\makecell{LLaVA-PLLuM\\12B-nc}} 
 & \rotatebox{60}{\makecell{LLaVA-Bielik\\11B-v2.6}} \\
\midrule
LLaVA-1.6-Mistral-7B & \textbf{57.38} & \textbf{58.68} & \textbf{55.47} \\
LLaVA-1.6-Vicuna-13B  & 49.76 & \textbf{51.4} & 48.71 \\
PaliGemma2-10B & \textbf{64.83} & \textbf{65.28} & \textbf{62.39} \\
Pixtral-12B  & 47.38 & 49.29 & 46.72 \\
Qwen2.5-VL-7B & 46.69 & 48.75 & 45.7 \\
\bottomrule
\end{tabular}
}
\caption{Preference rate (\%) of our models over baseline models judged by VLM (llava-onevision-qwen2-7b-si-hf) on XM3600 dataset for \textit{content description quality}.}
\label{tab:xm3600_results_vlm}
\end{table}

Table~\ref{tab:mmbehcnresults} presents the results on Polish and English variants of MMBench. The general drop in scores when switching from English to Polish confirms that Polish remains a challenging language for VLMs even within the same questions. Despite this, our \textsc{LLaVA-PLLuM-12b-nc} model outperforms the best LLaVA-1.6 baseline, achieving a $+9.55$ percentage point gain on the Polish split while maintaining similar performance in English. We interpret this gap as a considerable improvement in language understanding, distinct from minor fluctuations of $2$--$3$ points which are often negligible. Furthermore, our model surpasses strong open-source, competitors like \textsc{Qwen2.5-VL-7B} and \textsc{PaliGemma2-10B} on the Polish benchmark. Although \textsc{Pixtral-12B} achieves the highest score, its exact training data is undisclosed, making our transparent approach a valuable alternative for multilingual research. Moreover, Pixtral was trained using full fine-tuning, whereas our model relies on parameter-efficient LoRA adaptation.

The linguistic quality evaluation on XM3600 dataset summarized in Table~\ref{tab:xm3600_results_llm} shows that our models consistently outperform LLaVA-1.6-Mistral-7B, LLaVA-1.6-Vicuna-13B, and PaliGemma2-10B. However, they still lag behind Pixtral-12B and Qwen2.5-VL-7B. Our best-performing model, LLaVA-PLLuM-12b-nc, achieves a win rate comparable to Pixtral-12B (48\%), which motivated a more in-depth comparison with these stronger baselines using both human evaluation and a larger Claude Sonnet model as the judge. The results for content description quality are presented in Table~\ref{tab:xm3600_results_vlm}, where LLaVA-PLLuM-12b-nc again outperforms LLaVA-1.6-Mistral-7B, LLaVA-1.6-Vicuna-13B, and PaliGemma2-10B, while achieving performance comparable to Pixtral-12B and Qwen2.5-VL-7B, with win rates of 49.29\% and 48.75\%, respectively.

\begin{figure}[ht]
    \centering
    \includegraphics[width=1\linewidth]{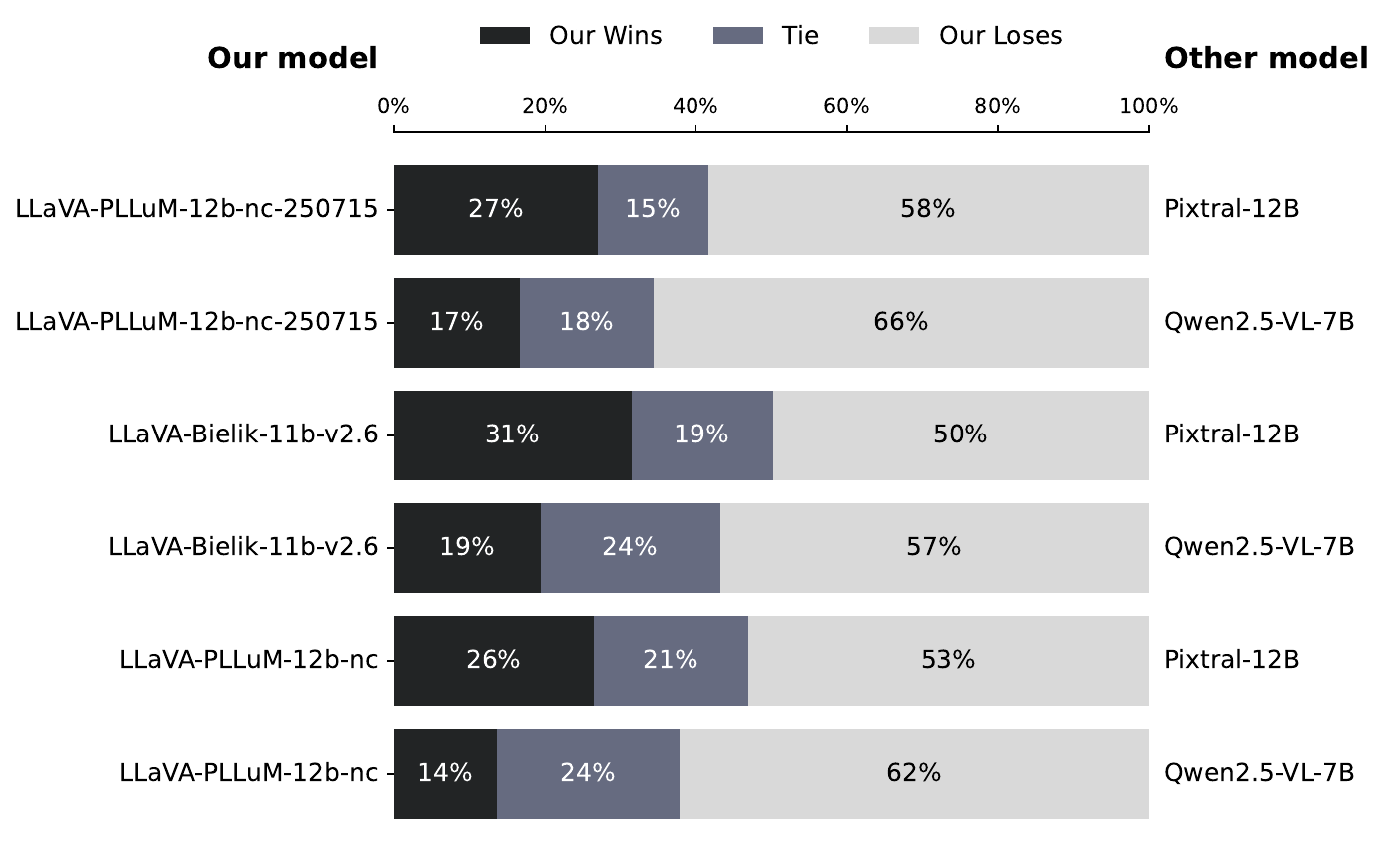}
    \caption{Automatic evaluation on a subset of the XM3600 dataset for the \textit{content description quality} criterion, using Claude Sonnet~4.5 as the judge.}
    \label{fig:xm3600_results_claude_content_eval}
\end{figure}

\begin{figure}[ht]
    \centering
    \includegraphics[width=1\linewidth]{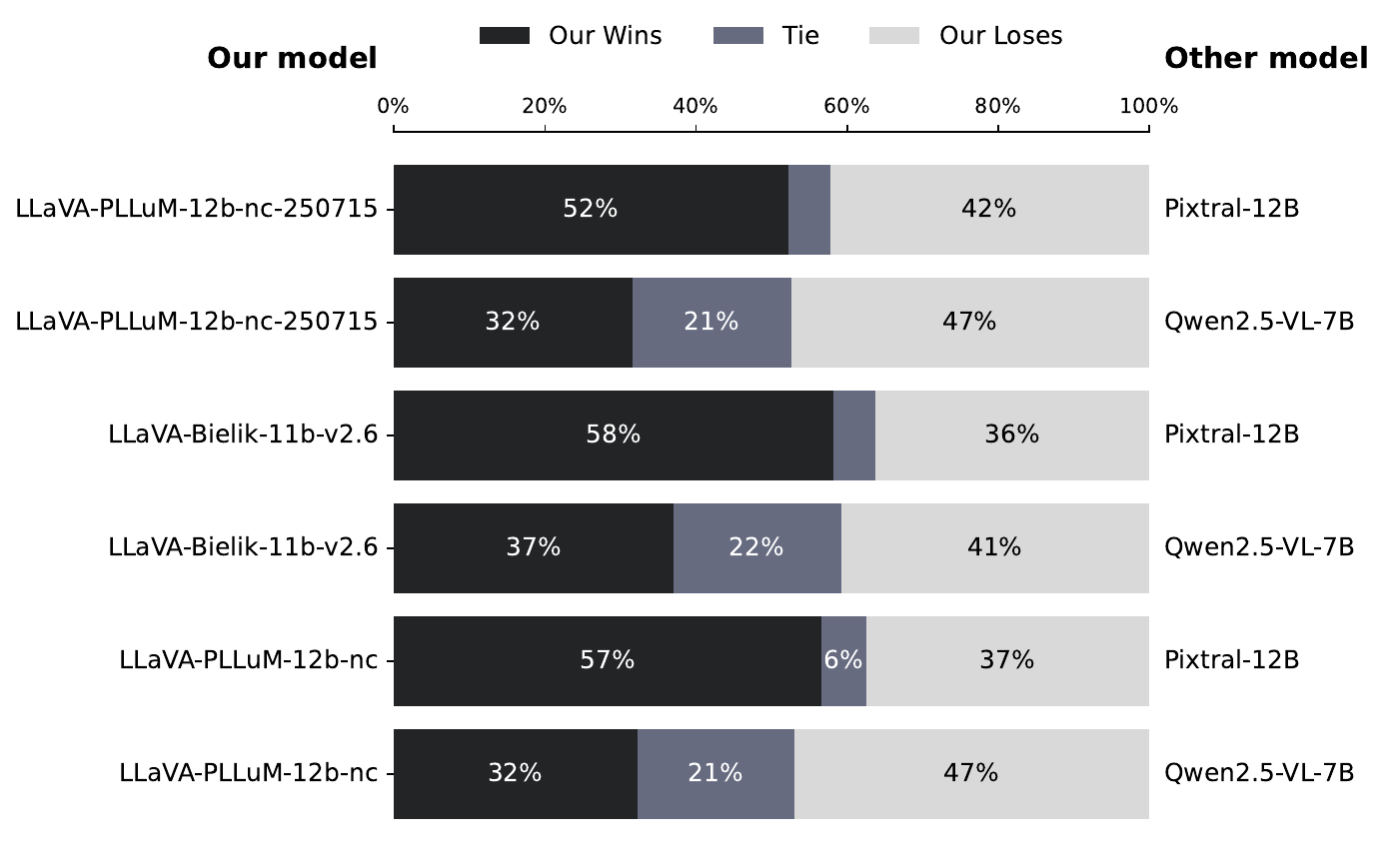}
    \caption{Automatic evaluation on a subset of the XM3600 dataset for the \textit{linguistic correctness} criterion, using Claude Sonnet~4.5 as the judge.}
    \label{fig:xm3600_results_claude_lang_eval}
\end{figure}

To further analyze these findings, we conduct an additional evaluation using Claude Sonnet as a judge.
In Figure~\ref{fig:xm3600_results_claude_lang_eval}, our models achieve higher win-tie rates against Pixtral~12B, indicating stronger linguistic fluency, but still underperform compared to the Qwen model. Notably, LLaVA-Bielik-11b-v2.6 achieves performance comparable to Qwen2.5-VL-7B in terms of linguistic quality, achieving a win-tie rate of 48.1\%. In contrast, Figure~\ref{fig:xm3600_results_claude_content_eval} presents a less favorable pattern for content description, where competing models generally outperform ours.

\begin{figure}
    \centering
    \includegraphics[width=1\linewidth]{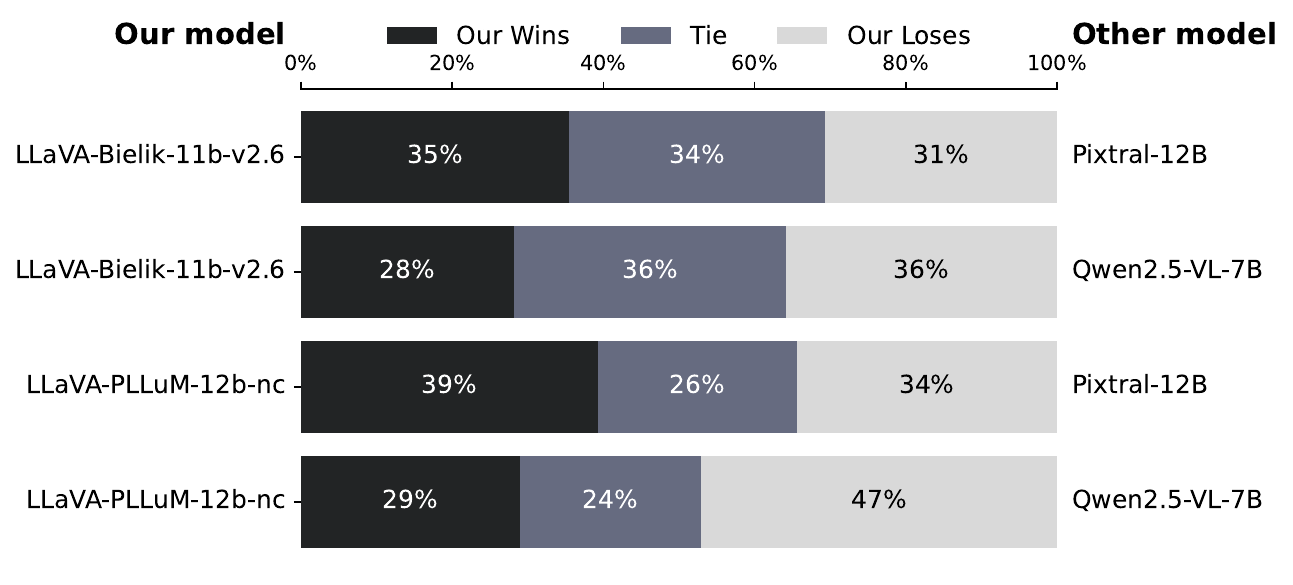}
    \caption{Human evaluation of \textit{image description quality} on a subset of the XM3600 dataset.}
    \label{fig:xm3600_results_human_content_eval}
\end{figure}

\begin{figure}
    \centering
    \includegraphics[width=1\linewidth]{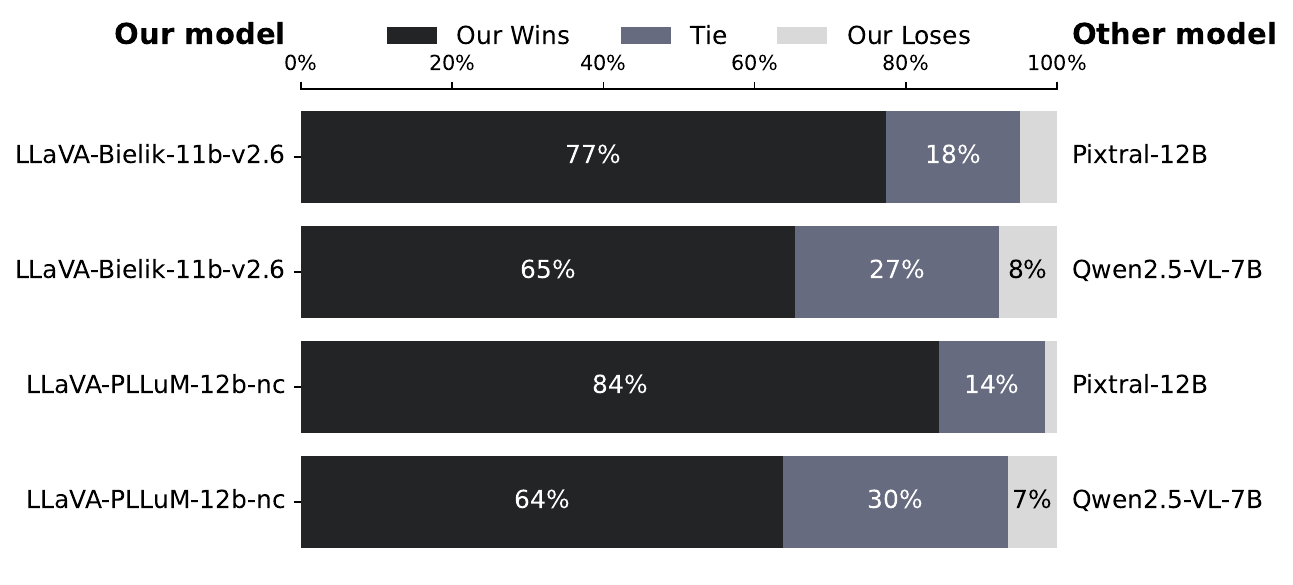}
    \caption{Human evaluation of \textit{linguistic correctness} quality on a subset of the XM3600 dataset.}
    \label{fig:xm3600_results_human_lang_eval}
\end{figure}

Figures \ref{fig:xm3600_results_human_content_eval} and \ref{fig:xm3600_results_human_lang_eval} present results of human evaluaton on the subset of XM3600 dataset (details of the number of samples are available in Appendix~\ref{app:manaul_numer_of_samples}). In contrast to the automatic preference judgments produced by Claude Sonnet, the human evaluation reveals a different performance profile. In particular, for linguistic correctness, manual assessment shows a strong advantage of our models over the baselines, with win rates of at least 64\% for all evaluated models and up to 84\% for the best-performing model, LLaVA-PLLuM-12b-nc, when compared against Pixtral-12B. This discrepancy between human and automatic evaluations suggests that fine-grained linguistic phenomena remain challenging for LLM-based judges to assess reliably. Moreover, the manual evaluation indicates that for the content description criterion our models still underperform compared to Qwen, while they achieve an advantage over Pixtral — most notably, the best-performing model, LLaVA-PLLuM-12b-nc, attains a 39\% win rate versus 34\% for Pixtral — a trend that was considerably less apparent in the automatic evaluation.

\subsection{Qualitative analysis}

Next, we present two qualitative examples in Figure~\ref{fig:qualitive_exaples} that illustrate the performance gains of our VLMs. In the first example, the models are shown an image of cat inside of a dishwasher and are asked, ``Where is the cat?'' Our model provides a correct response in terms of both content and linguistic accuracy. In contrast, Qwen2.5-VL-7B-Instruct states that the cat is inside a ``washing machine for dishes,'' which is not a correct description of the image. 
The most original answer belongs to Llava-v1.6-vicuna-13b-hf that presents an erroneous answer in both content and language: the model, creatively, introduces non-standard and invented terminology.
In the second example, the models are presented with a map of Poland depicting its neighbouring countries and are asked the question, ``How many neighbors does Poland have?'' LLaVA-PLLuM-12b-nc (\textit{our model}) produces a correct response, stating that Poland has seven neighboring countries and correctly listing all of them. In contrast, Qwen2.5-VL-7B-Instruct provides an incorrect answer because it omits Slovakia and Russia from the list and uses incorrect Polish syntax.
Pixtral-12B-2409 produces an incorrect response in both content and language: it fails to mention Germany and Belarus, lists Slovakia twice, and contains multiple linguistic errors.
Overall, these results demonstrate an improvement in both content recognition and language accuracy achieved by our model when compared to the other evaluated models. We provide further examples in Appendix~\ref{app:internal}.

\section{Conclusion and Future Work}

We presented a practical approach for building vision-language models for a non-English, morphologically rich language using large-scale automated translation. 
By adapting the LLaVA-NeXT training recipe to Polish and relying mainly on translated multimodal data, complemented with limited synthetic data, we obtained consistent improvements over English-centric baselines on a Polish-adapted MMBench and in human evaluations of caption quality. 
These results indicate that automatic translation, combined with basic filtering, can be an effective way to bootstrap multimodal models for languages with limited native resources.

Future work may extend this approach to improve cultural coverage through more language-specific or region-specific data. 
In addition, more advanced data filtering and quality control methods could further reduce translation noise. 

\section*{Limitations}
This study has several limitations. Firstly, the quality of the training data relies heavily on automatic translation, which may introduce translationese, subtle semantic drift and unnatural phrasing that could affect the behaviour of the model. Although COMET-based filtering was applied, no systematic ablations were conducted on translation-quality thresholds, and the downstream impact of translation-induced artefacts was not directly measured.

Secondly, although the design goals emphasise Polish OCR capabilities and Poland-specific knowledge, the reported evaluations primarily focus on general multimodal benchmarks. As we did not include a dedicated OCR benchmark or a Poland-specific knowledge evaluation that isolates these target capabilities, our ability to directly validate these stated objectives remains limited.

Third, cultural and contextual coverage is limited. Although Polish-language supervision was expanded through translation and Wikipedia-based resources, most visual tasks and source datasets originate from English-centric benchmarks and only partially reflect Polish contexts. Consequently, model performance in authentic Polish scenarios may not be fully captured. Furthermore, the evaluation is restricted to a limited set of benchmarks and metrics that focus mainly on accuracy and linguistic correctness. Other aspects, such as deeper reasoning, factual grounding, robustness to cultural nuances and real-world usability, remain underexplored.

Furthermore, comparisons across evaluation modes are not fully aligned. In the XM3600 automatic-judge setup, ties are not permitted, whereas human and Claude-based evaluations allow them. Consequently, preference rates cannot be measured on a shared scale, which makes direct comparison across evaluation methods difficult.

Finally, several key design choices were not subjected to comprehensive ablations, including the use of SigLIP2 versus CLIP visual encoders, the Polish-to-English training ratio, the benefits of the two-stage training procedure and the use of LoRA versus full language-model fine-tuning. Future work should systematically evaluate these factors to better understand their individual contributions and interactions.

\section*{Acknowledgements}

We acknowledge Polish high-performance computing infrastructure PLGrid (HPC Centers: ACK Cyfronet AGH) for providing computer facilities and support within computational grant no. PLG/2025/018129.
This work was also supported by the Ministry of Digital Affairs (subsidy no. 8/WII/DBI/2025).

\bibliography{custom}

\clearpage

\appendix

\section{Traning Details}
\label{app:training-details}

Table~\ref{tab:training} lists the set of hyperparameters and computational resources used for both training stages. The configuration is organized into two distinct columns, corresponding to the pre-training and instruction-tuning phases. It details specific settings for batch sizes, context lengths, and learning rates across different model components, as well as the LoRA adapter configuration and hardware infrastructure employed.

\begin{table}[t]
\centering
\small
\renewcommand{\arraystretch}{1.1}
\begin{tabularx}{\linewidth}{@{}lYY@{}}
\toprule
    & \textbf{Stage 1} & \textbf{Stage 2} \\
\midrule

\tblsec{Training data}
Dataset     & Section \ref{pretrain-data}   & Section \ref{instruct-data} \\
\# Samples  & 595K                          & 2.0M \\

\tblsec{Model \& Optimization}
Trainable Parameters 
& Projector 
& Full Model 
\\
Context (tokens) & 8{,}192       & 8{,}192 \\
Batch size  & 256   & 128   \\
LR (Vision) & -- & $2\times10^{-6}$ \\
LR (Projector) & $1\times10^{-3}$ & $2\times10^{-5}$ \\
LR (LLM) & -- & $2\times10^{-5}$ \\
LoRA $r$ & -- & 128 \\
LoRA $\alpha$ & -- & 256 \\
LoRA Dropout & -- & 0.05 \\
GPU hours           & 336 \newline \scriptsize{(A100 40GB)} & 1{,}344 \newline \scriptsize{(GH200 96GB)} \\
\bottomrule

\end{tabularx}
\caption{Training configuration of the model. Dash indicates that the component is frozen.}
\label{tab:training}
\end{table}

\section{Manual Evaluation}\label{app:manaul_numer_of_samples}

Table~\ref{tab:manual_eval_numer_samples} reports the number of samples manually evaluated by human annotators. These samples are a subset of the 500 observations used for the automatic evaluation conducted with Claude Sonnet~4.5. 
The evaluation was conducted in an anonymous setting, where annotators were not informed about the identity of the models generating the responses. This design was intended to prevent bias and ensure a fair and objective comparison of model outputs.

\begin{table}[h]
   \centering
    \small
    \resizebox{\columnwidth}{!}{
        \begin{tabular}{ccc}
            \toprule

 Our model & Other model & N \\
\midrule
 LLaVA-Bielik-11b-v2.6 & Pixtral-12B & 62
\\
LLaVA-Bielik-11b-v2.6 & Qwen2.5-VL-7B & 78 \\
LLaVA-PLLuM-12b-nc  & Pixtral-12B & 122\\
LLaVA-PLLuM-12b-nc & Qwen2.5-VL-7B & 138\\
\bottomrule
\end{tabular}
}
\caption{Number of sample pairs assessed in manual evaluation.}
\label{tab:manual_eval_numer_samples}
\end{table}

\section{Prediction Examples} \label{app:internal}

\begin{figure}[H]
    \centering
    \includegraphics[width=1\linewidth]{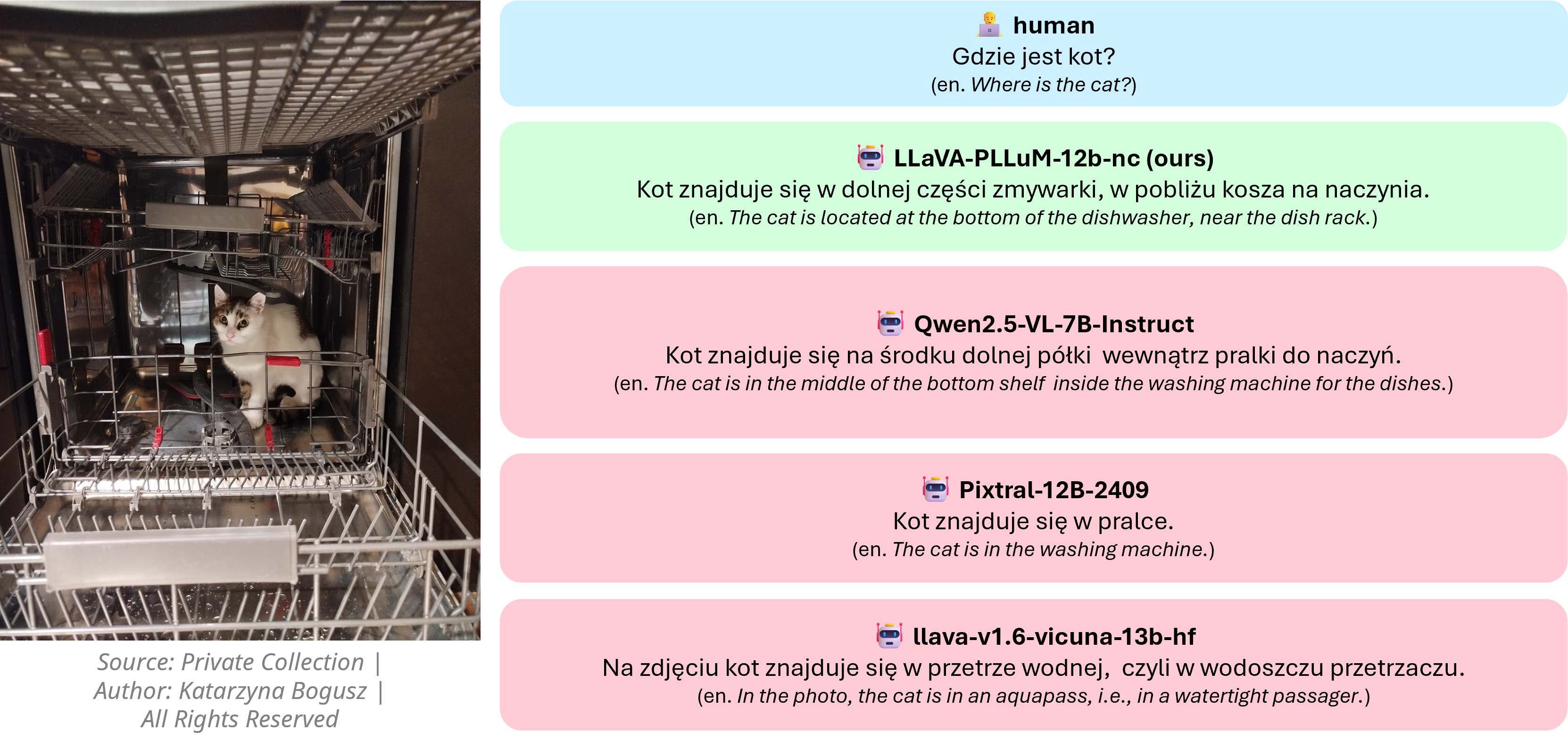}
    \caption{LLaVA-PLLuM-12B-nc provided correct response. Qwen2.5-VL-7B and Pixtral-12B provided an incorrect location, while LLaVA-1.6-Vicuna-13B created a fictional location.}
    \label{fig:slide1a}
\end{figure}

\begin{figure}[H]
    \centering
    \includegraphics[width=1\linewidth]{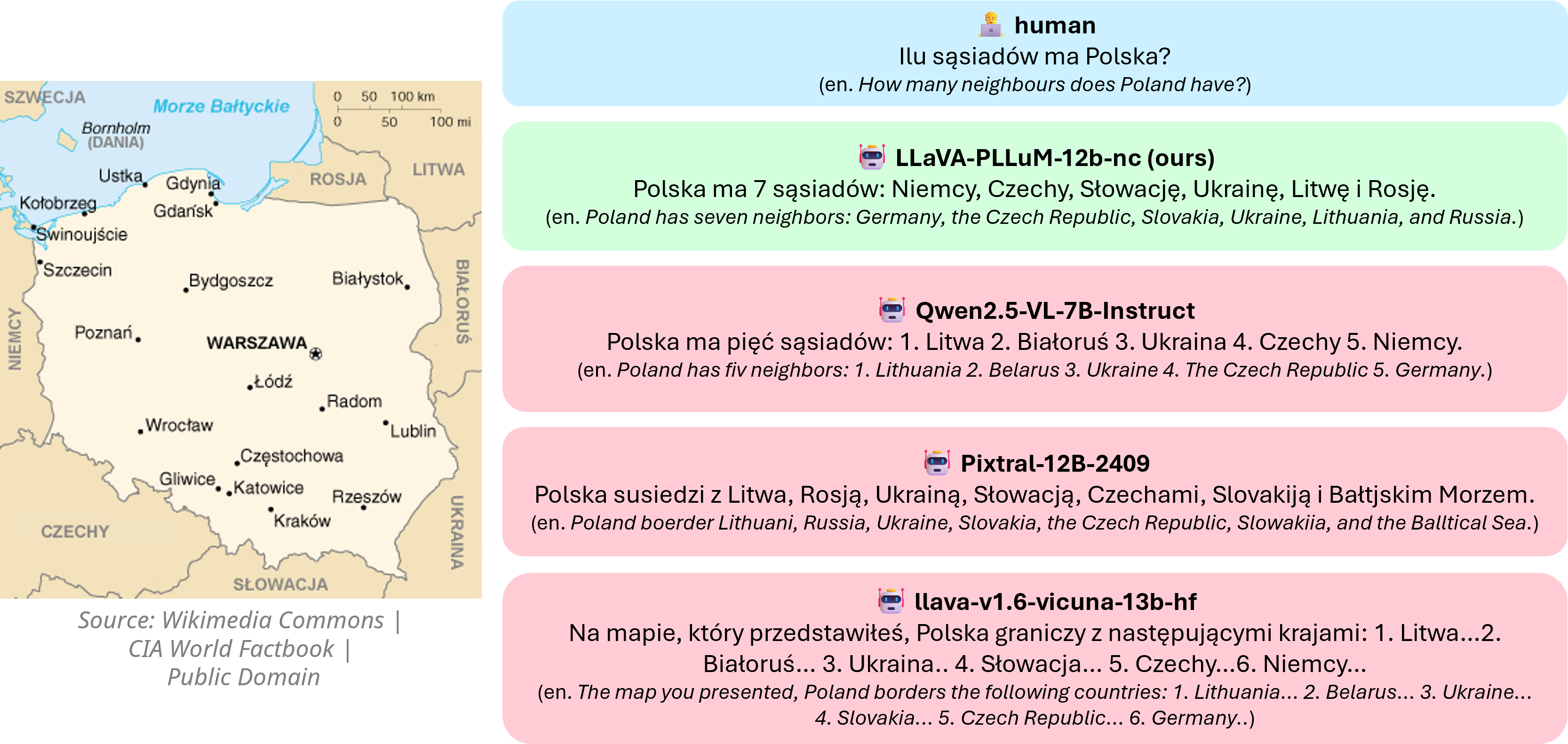}
    \caption{LLaVA-PLLuM-12B-nc provided correct response. Qwen2.5-VL-7B listed an incorrect number of neighbors, while Pixtral-12B and LLaVA-1.6-Vicuna-13B generated linguistic errors and factual inaccuracies.}
    \label{fig:slide1b}
\end{figure}

\begin{figure}[H]
    \centering
    \includegraphics[width=1\linewidth]{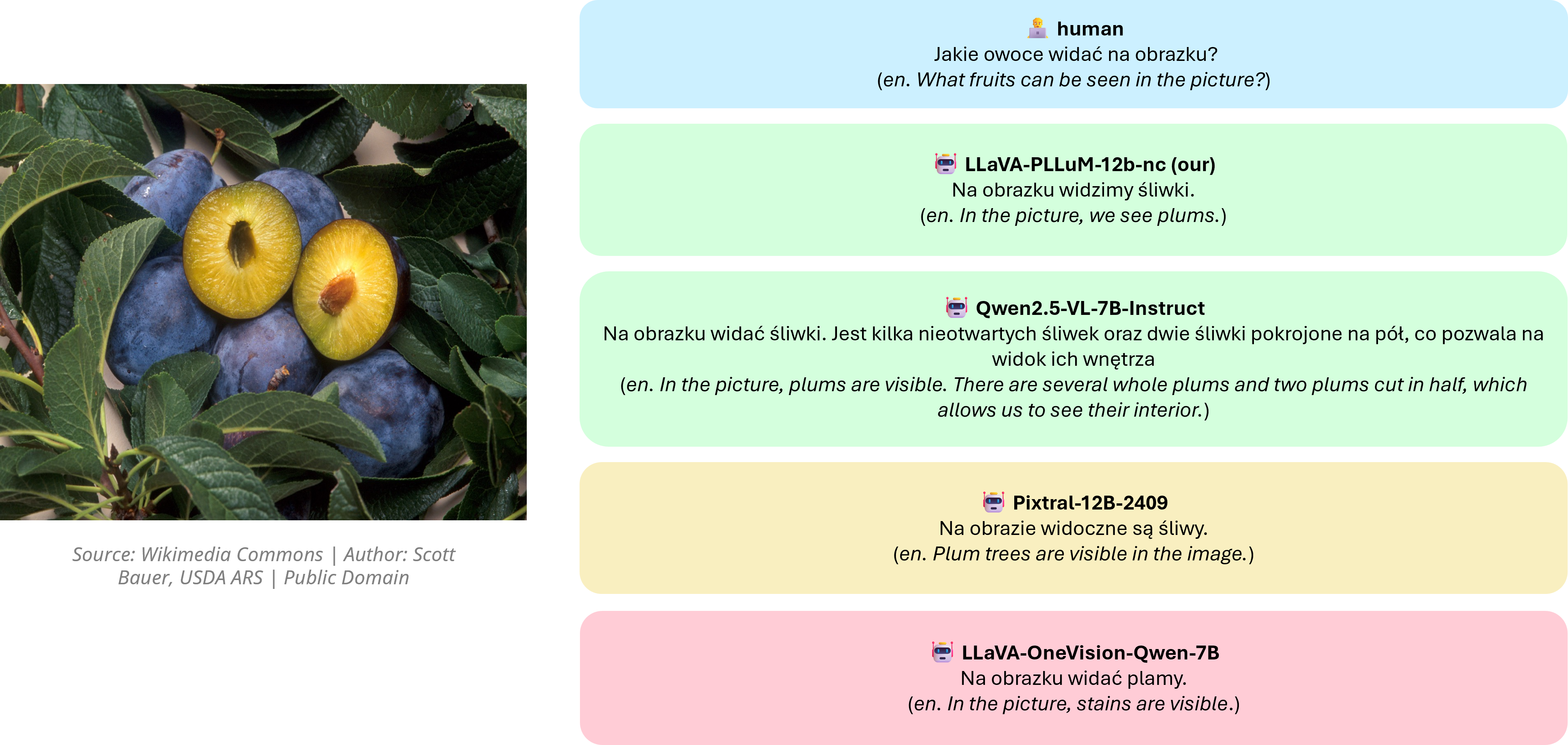}
    \caption{LLaVA-PLLuM-12B-nc and Qwen2.5-VL-7B provided correct responses. Pixtral-12B used the name of the tree instead of the fruit while LLaVA-1.6-Vicuna-13B made up a fictional fruit.}
    \label{fig:slide2}
\end{figure}

\begin{figure}[H]
    \centering
    \includegraphics[width=1\linewidth]{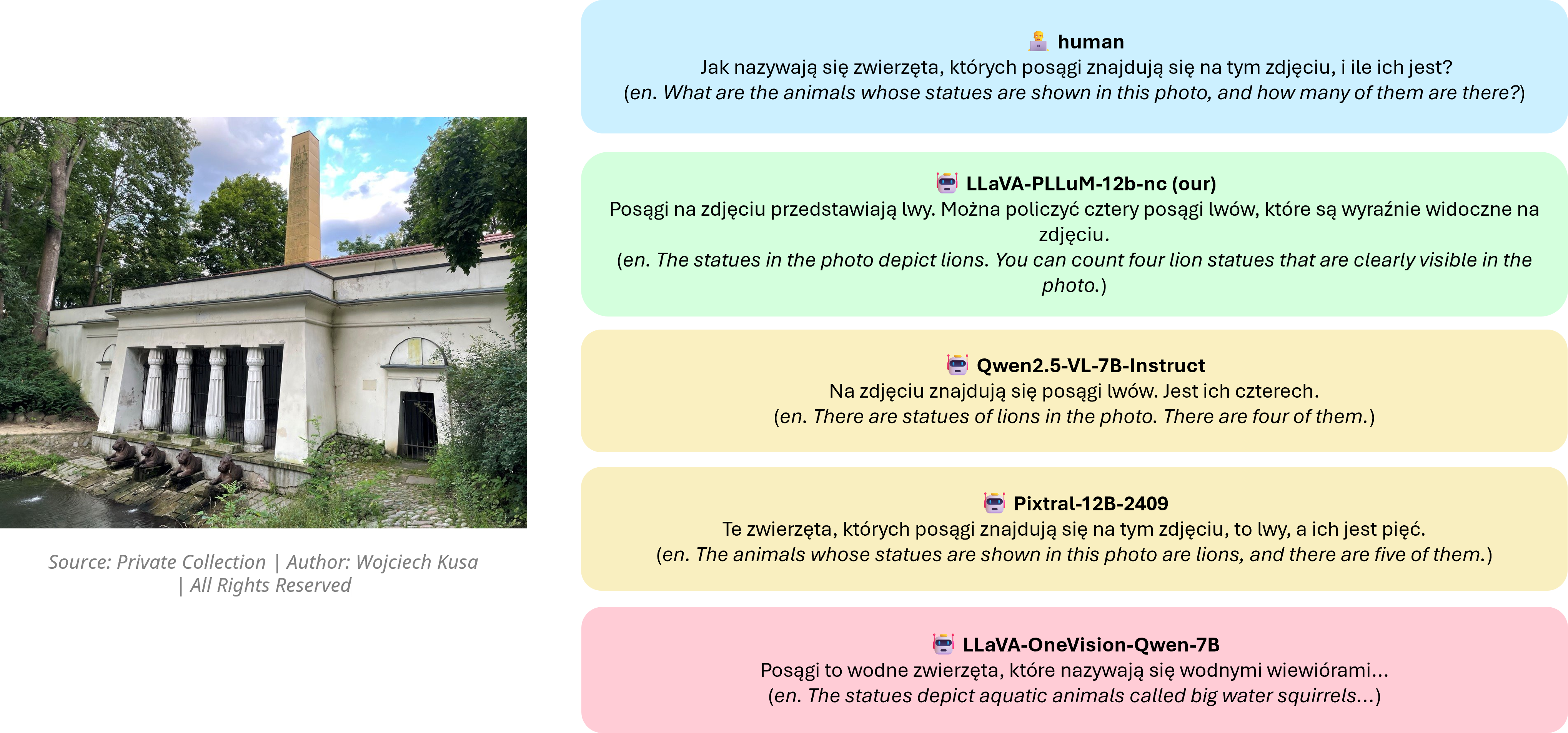}
    \caption{Qwen2.5-VL-7B correctly identified the object and its quantity, but used an incorrect grammatical inflection. Pixtral-12B provided the correct label but an incorrect quantity. LLaVA-1.6-Vicuna-13B generated a non-existent animal.}
    \label{fig:slide3}
\end{figure}

\begin{figure}[H]
    \centering
    \includegraphics[width=1\linewidth]{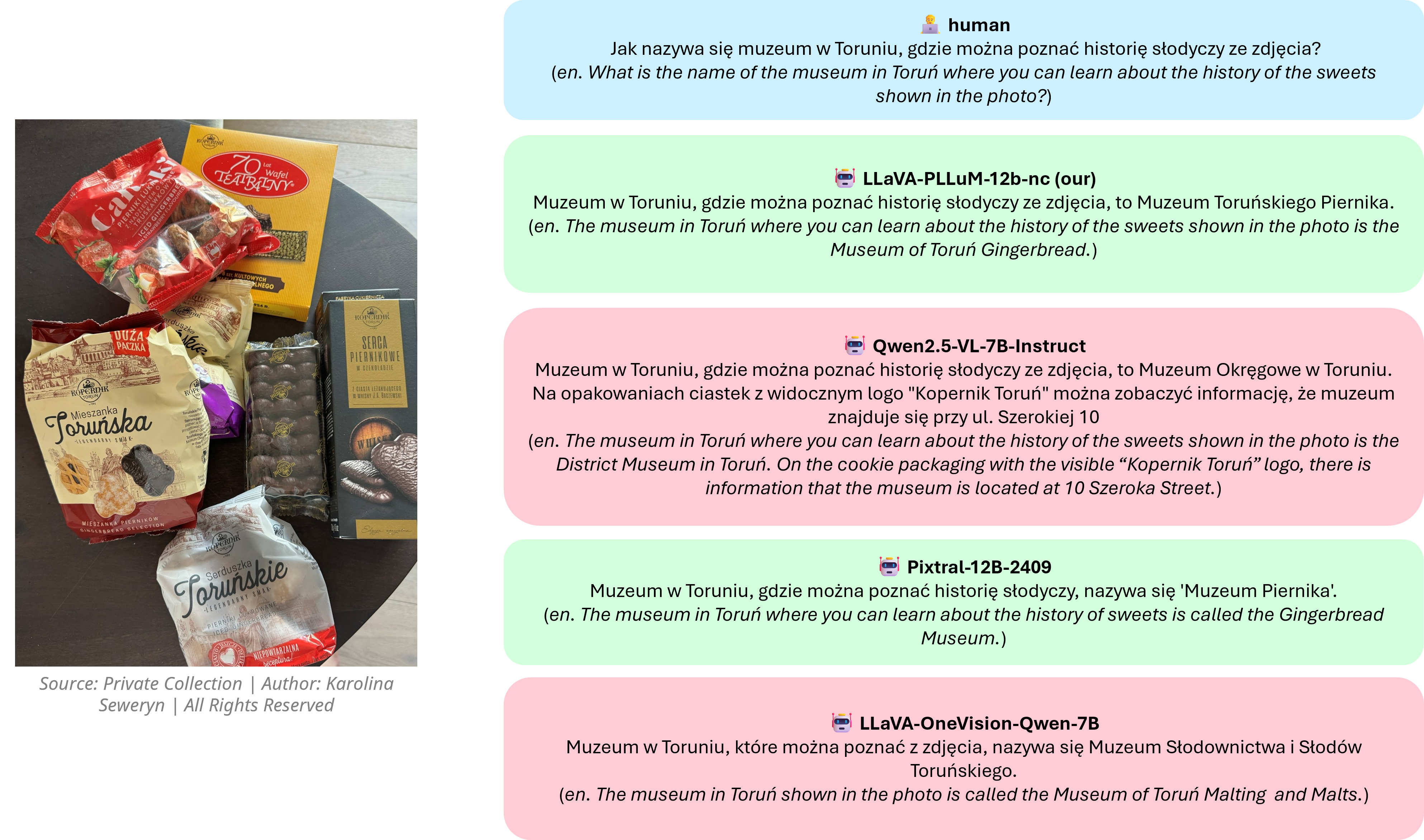}
    \caption{LLaVA-PLLuM-12B-nc and Pixtral-12B correctly identified the museum, whereas Qwen2.5-VL-7B and LLaVA-1.6-Vicuna-13B provided an incorrect museum name.}
    \label{fig:slide4}
\end{figure}

To evaluate the models’ ability to capture and interpret the Polish cultural context, we curated and annotated a small image dataset and assessed the models’ responses to these samples. Figures~\ref{fig:slide1a}-\ref{fig:slide4} present a few illustrative examples.

\section{Discussion of MMbench issues} \label{app:mmbench}

\begin{table}[ht]
\centering
\footnotesize
{\renewcommand{\arraystretch}{0.9}
\begin{tabularx}{\columnwidth}{l X c}
\toprule
Cat. & Problem & N \\
\midrule
1a & More than one answer is correct & 8 \\
1b & None of the answers is correct & 4 \\
1c & Answer marked as correct is incorrect & 4 \\
1d & The content of the picture is ambiguous & 3 \\
1e & It is not possible to predict the consequences of the action & 5 \\
1f & Unfortunate phrasing that might result in biased or stereotypical interpretation & 3 \\
1g & The correct answer contains minor mistakes & 8 \\
1h & Necessary context missing & 6 \\
1i & The relation between the picture and the question/answer is flawed & 7 \\
\bottomrule
\end{tabularx}
}
\caption{Categorization of MMBench issues. N refers to number of questions within a category out a total of 1292 unique questions.}
\label{tab:MMbench-problems}
\end{table}

Table~\ref{tab:MMbench-problems} presents various categories of MMBench problems we identified. While categories 1a-1c  are largely self-explanatory, the remaining categories require further clarification. Category 1d includes cases in which it is difficult to identify the depicted object or its relevant characteristics. Category 1e refers to questions that prompt the model to predict future events based on the images; however, in these cases, the consequences of the depicted action are not inferable from the image alone. Category 1f comprises cases in which the formulation of the question in relation to the image can lead to a stereotypical interpretations (e.g., asking “What is the color of this object?” when referring to a photo of a Black man wearing a purple t-shirt). The category 1g encloses cases in which – in contrast to the category 1b – it is possible to indicate the most plausible answer; however, this answer contains minor factual inaccuracies. The category 1h includes questions for which essential information is missing, rendering them unanswerable. Lastly, the category 1i refers to cases in which the logical relationship between the image and the question or the answers is flawed. Examples of all categories are provided in the annex. Although some categories may be considered overlapping, each question was assigned to just one category based on its prevailing characteristic. 

In total, 3.56\% of the unique questions in MMBench were identified as inaccurate or otherwise flawed. 
In addition, a subset of questions was identified as being rooted in a foreign linguistic or cultural context. However, it should be noted that these questions are not substantively flawed; instead, they require specific knowledge that is not central for a VLM whose primary objective is to understand and reflect Polish linguistic and cultural context, thereby posing localization challenges. Two such categories are presented in Table~\ref{tab:foreign-context}. 

\begin{table}[h!]
\centering
\footnotesize
\setlength{\tabcolsep}{2pt}
\renewcommand{\arraystretch}{0.85}
\begin{tabular}{@{}llr@{}}
\toprule
\textbf{Cat.} & \textbf{Challenge} & \textbf{\#Q} \\
\midrule
2a & Phrases in Chinese/Japanese & 9 \\
2b & Identifying non-European people/buildings/dishes & 30 \\
\bottomrule
\end{tabular}
\caption{Questions rooted in a foreign context.}\label{tab:foreign-context}
\end{table}

\begin{figure}[H]
  \centering
  \begin{minipage}{0.35\linewidth}
    \centering
    \includegraphics[width=\linewidth]{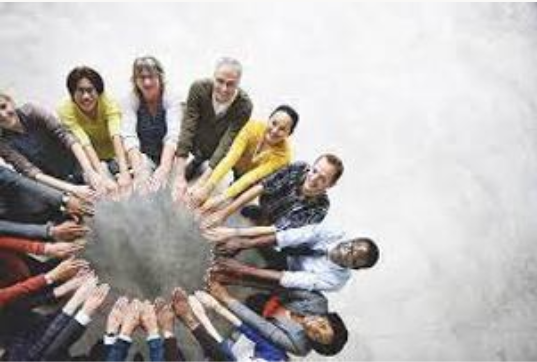}
  \end{minipage}
  \hfill
  \begin{minipage}{0.6\linewidth}
    \small
    \setlength{\parskip}{0.25em}
    
    \textbf{Question:}
If you were to join the group shown in the image, which role would you most likely assume?

    \begin{enumerate}[label=\Alph*., leftmargin=*, itemsep=0.15em]
      \item The facilitator of the meeting
      \item A group member
      \item The note-taker or observer
      \item A presenter or speaker
    \end{enumerate}

    \textbf{Original Answer:} B
    
    \textbf{Explanation:} It is possible to assume multiple roles in this group. 
\end{minipage}
  \caption{MMBench example from Category 1a: More than one answer is correct.}
  \label{fig:cat-1a-1}
\end{figure}

\begin{figure}[H]
  \centering
  \begin{minipage}{0.35\linewidth}
    \centering
    \includegraphics[width=\linewidth]{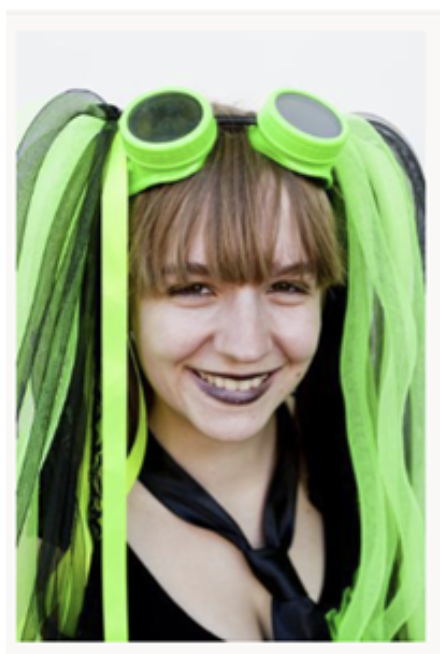}
  \end{minipage}
  \hfill
  \begin{minipage}{0.6\linewidth}
    \small
    \setlength{\parskip}{0.25em}
    
    \textbf{Question:}
    Based on the image, which aspects of the woman's appearance contribute to the
    impression of playfulness?
    
    \begin{enumerate}[label=\Alph*., leftmargin=*, itemsep=0.15em]
      \item The green hair and goggles
      \item The tie
      \item Her unconventional style
      \item Her engaging smile
    \end{enumerate}

    \textbf{Original Answer:} A
    
    \textbf{Explanation:} All answers can be considered correct.
  \end{minipage}
  \caption{MMBench example from Category 1a: More than one answer is correct.}
  \label{fig:cat-1a-2}
\end{figure}

\begin{figure}[H]
  \centering
  \begin{minipage}{0.35\linewidth}
    \centering
    \includegraphics[width=\linewidth]{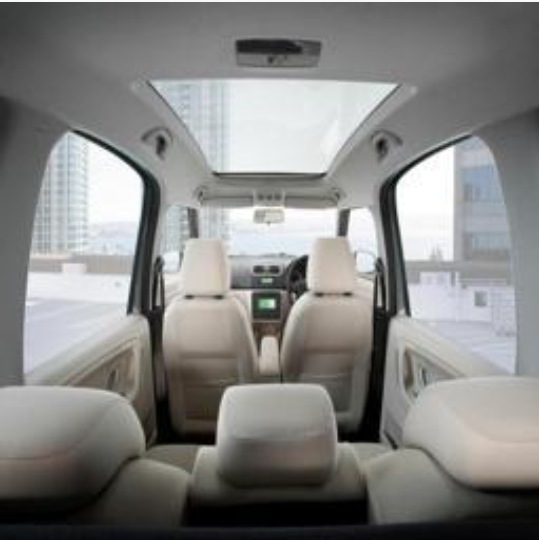}
  \end{minipage}
  \hfill
  \begin{minipage}{0.6\linewidth}
    \small
    \setlength{\parskip}{0.25em}
    
    \textbf{Question:}
    Which scene category matches this image the best? 
    
    \begin{enumerate}[label=\Alph*., leftmargin=*, itemsep=0.15em]
      \item bowling\_alley
      \item airplane\_cabin
      \item porch
      \item shed
    \end{enumerate}

    \textbf{Original Answer:} B
    
    \textbf{Explanation:} None of the answer seems correct. The image shows probably the inside of a car.
  \end{minipage}
  \caption{MMBench example from Category 1b: None of the answers is correct.}
  \label{fig:cat-1b-1}
\end{figure}

\begin{figure}[H]
  \centering
  \begin{minipage}{0.35\linewidth}
    \centering
    \includegraphics[width=\linewidth]{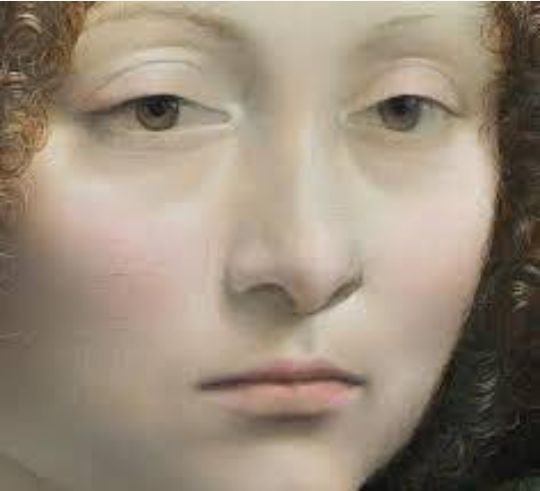}
  \end{minipage}
  \hfill
  \begin{minipage}{0.6\linewidth}
    \small
    \setlength{\parskip}{0.25em}
    
    \textbf{Question:}
    The image presents an abstract form that could be interpreted in multiple ways. This ambiguity is a characteristic of:
    
    \begin{enumerate}[label=\Alph*., leftmargin=*, itemsep=0.15em]
      \item Constructivism
      \item Futurism
      \item Suprematism
      \item Minimalism
    \end{enumerate}

    \textbf{Original Answer:} C
    
    \textbf{Explanation:} This painting is neither an abstract form nor does it belong to any of the art movements listed as answers.
  \end{minipage}
  \caption{MMBench example from Category 1b: None of the answers is correct.}
  \label{fig:cat-1b-2}
\end{figure}

\begin{figure}[H]
  \centering
  \begin{minipage}{0.35\linewidth}
    \centering
    \includegraphics[width=\linewidth]{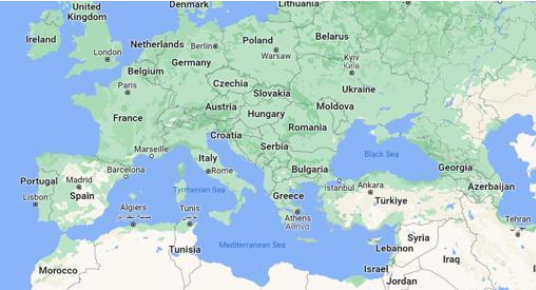}
  \end{minipage}
  \hfill
  \begin{minipage}{0.6\linewidth}
    \small
    \setlength{\parskip}{0.25em}
    
    \textbf{Question:}
    What direction is Ukraine in the Black Sea?

    \begin{enumerate}[label=\Alph*., leftmargin=*, itemsep=0.15em]
      \item east
      \item south
      \item west
      \item north
    \end{enumerate}

    \textbf{Original Answer:} A
    
    \textbf{Explanation:} Ukraine is north from the Black Sea, not east.
  \end{minipage}
  \caption{MMBench example from Category 1c: Answer marked as correct is incorrect.}
  \label{fig:cat-1c-1}
\end{figure}

\begin{figure}[H]
  \centering
  \begin{minipage}{0.35\linewidth}
    \centering
    \includegraphics[width=\linewidth]{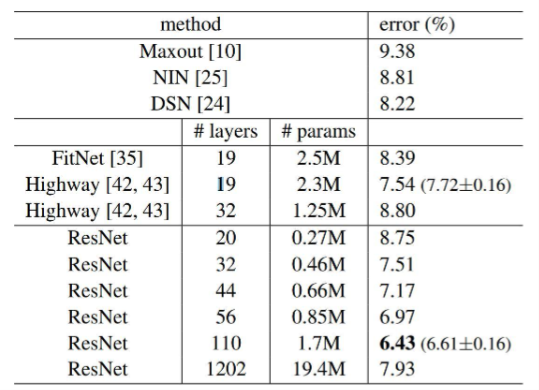}
  \end{minipage}
  \hfill
  \begin{minipage}{0.6\linewidth}
    \small
    \setlength{\parskip}{0.25em}
    
    \textbf{Question:}
What is the error of DSN?

    \begin{enumerate}[label=\Alph*., leftmargin=*, itemsep=0.15em]
      \item 7.96
      \item 9.38
      \item 8.81
      \item 8.22
    \end{enumerate}

    \textbf{Original Answer:} C
    
    \textbf{Explanation:} The DSN error in the image is 8.22, not 8.81. 
\end{minipage}
  \caption{MMBench example from Category 1c: Answer marked as correct is incorrect.}
  \label{fig:cat-1c-2}
\end{figure}

\begin{figure}[H]
  \centering
  \begin{minipage}{0.35\linewidth}
    \centering
    \includegraphics[width=\linewidth]{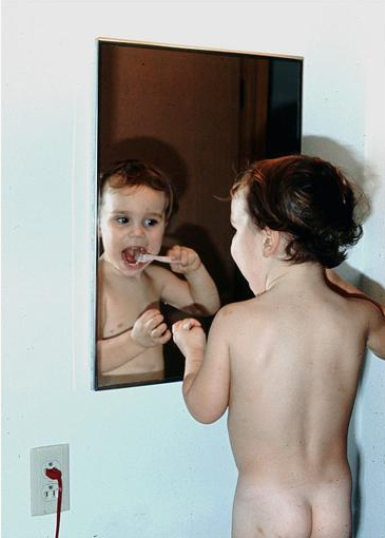}
  \end{minipage}
  \hfill
  \begin{minipage}{0.6\linewidth}
    \small
    \setlength{\parskip}{0.25em}
    
    \textbf{Question:}
Which is the main topic of the image

    \begin{enumerate}[label=\Alph*., leftmargin=*, itemsep=0.15em]
      \item A little boy brushing his teeth naked
      \item A little boy brushing his teeth with clothes on
      \item A little girl brushing her teeth naked
      \item A little boy taking a bath naked
    \end{enumerate}

    \textbf{Original Answer:} A
    
    \textbf{Explanation:} It is unclear whether the picture shows a little boy or a little girl.
\end{minipage}
  \caption{MMBench example from Category 1d: The content of the picture is ambiguous.}
  \label{fig:cat-1d-1}
\end{figure}

\begin{figure}[H]
  \centering
  \begin{minipage}{0.35\linewidth}
    \centering
    \includegraphics[width=\linewidth]{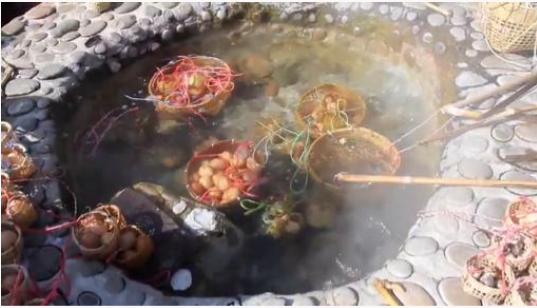}
  \end{minipage}
  \hfill
  \begin{minipage}{0.6\linewidth}
    \small
    \setlength{\parskip}{0.25em}
    
    \textbf{Question:}
Which action is performed in this image? 

    \begin{enumerate}[label=\Alph*., leftmargin=*, itemsep=0.15em]
      \item cooking egg
      \item barbequing
      \item cooking sausages
      \item cooking on campfire
    \end{enumerate}

    \textbf{Original Answer:} A
    
    \textbf{Explanation:} It is unclear what action is being performed in the picture.
\end{minipage}
  \caption{MMBench example from Category 1d: The content of the picture is ambiguous.}
  \label{fig:cat-1d-2}
\end{figure}

\begin{figure}[H]
  \centering
  \begin{minipage}{0.35\linewidth}
    \centering
    \includegraphics[width=\linewidth]{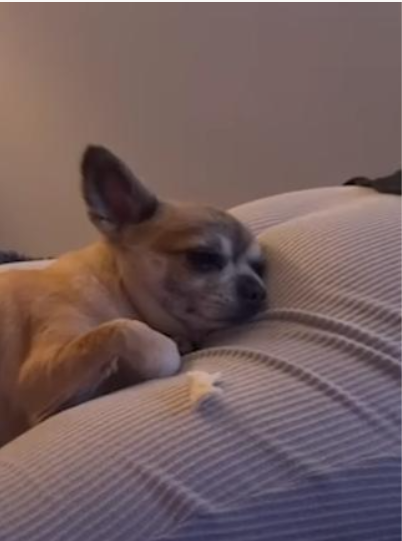}
  \end{minipage}
  \hfill
  \begin{minipage}{0.6\linewidth}
    \small
    \setlength{\parskip}{0.25em}
    
    \textbf{Question:}
What will happen next?

    \begin{enumerate}[label=\Alph*., leftmargin=*, itemsep=0.15em]
      \item the dog is gonna sleep
      \item the person is gonna fart on the dog
      \item the dog is gonna bite the person
      \item both A,B, and C
    \end{enumerate}

    \textbf{Original Answer:} A
    
    \textbf{Explanation:} The dog falling asleep is not the only possible outcome of the situation.
\end{minipage}
  \caption{MMBench example from Category 1e: It is not possible to predict the consequences of the action.}
  \label{fig:cat-1e-1}
\end{figure}

\begin{figure}[H]
  \centering
  \begin{minipage}{0.35\linewidth}
    \centering
    \includegraphics[width=\linewidth]{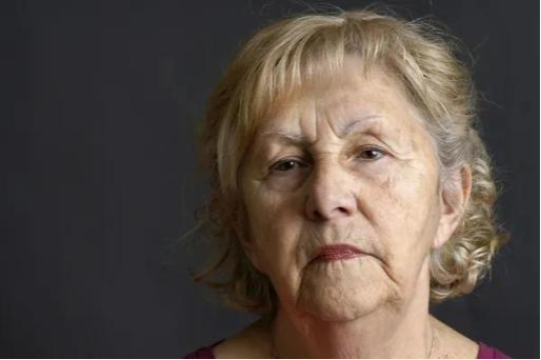}
  \end{minipage}
  \hfill
  \begin{minipage}{0.6\linewidth}
    \small
    \setlength{\parskip}{0.25em}
    
    \textbf{Question:}
What will happen next?

    \begin{enumerate}[label=\Alph*., leftmargin=*, itemsep=0.15em]
      \item this person is gonna laugh
      \item this person is gonna get mad
      \item this person is gonna cry
      \item both A,B, and C
    \end{enumerate}

    \textbf{Original Answer:} C
    
    \textbf{Explanation:} This person might cry, but she also might laugh, get mad, or stay indifferent.
\end{minipage}
  \caption{MMBench example from Category 1e: It is not possible to predict the consequences of the action.}
  \label{fig:cat-1e-2}
\end{figure}

\begin{figure}[H]
  \centering
  \begin{minipage}{0.35\linewidth}
    \centering
    \includegraphics[width=\linewidth]{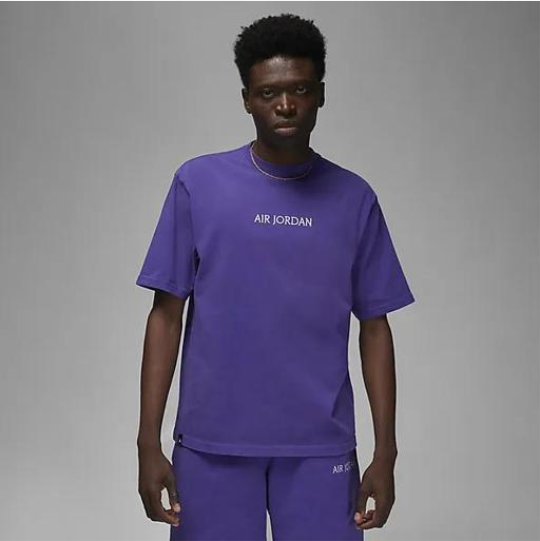}
  \end{minipage}
  \hfill
  \begin{minipage}{0.6\linewidth}
    \small
    \setlength{\parskip}{0.25em}
    
    \textbf{Question:}
what is the color of this object?

    \begin{enumerate}[label=\Alph*., leftmargin=*, itemsep=0.15em]
      \item purple
      \item pink
      \item gray
      \item orange
    \end{enumerate}

    \textbf{Original Answer:} A
    
    \textbf{Explanation:} This is an unfortunate phrasing that may result in biased or stereotypical interpretation. In the Polish version of the benchmark we changed the question to `What is the color of this man’s clothing?'.
\end{minipage}
  \caption{MMBench example from Category 1f: Unfortunate phrasing that might result in biased or stereotypical interpretation.}
  \label{fig:cat-1f-1}
\end{figure}

\begin{figure}[H]
  \centering
  \begin{minipage}{0.35\linewidth}
    \centering
    \includegraphics[width=\linewidth]{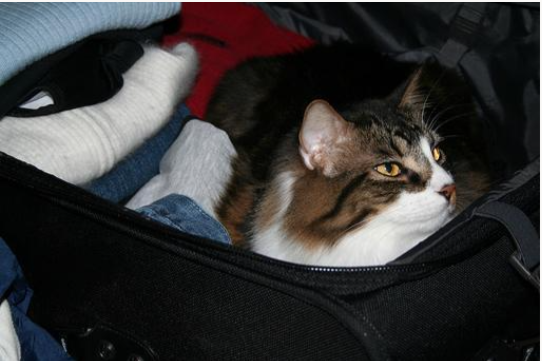}
  \end{minipage}
  \hfill
  \begin{minipage}{0.6\linewidth}
    \small
    \setlength{\parskip}{0.25em}
    
    \textbf{Question:}
Which one is the correct caption of this image?

    \begin{enumerate}[label=\Alph*., leftmargin=*, itemsep=0.15em]
      \item A small tower that has a clock at the top.
      \item A furry cat sleeping inside a packed suitcase
      \item A white bathroom sink sitting next to a walk in shower.
      \item a dog in a field with a frisbee in its mouth
    \end{enumerate}

    \textbf{Original Answer:} B
    
    \textbf{Explanation:} The cat is not sleeping.
\end{minipage}
  \caption{MMBench example from Category 1g: The correct answer contains minor mistakes. }
  \label{fig:cat-1g-1}
\end{figure}

\begin{figure}[H]
  \centering
  \begin{minipage}{0.35\linewidth}
    \centering
    \includegraphics[width=\linewidth]{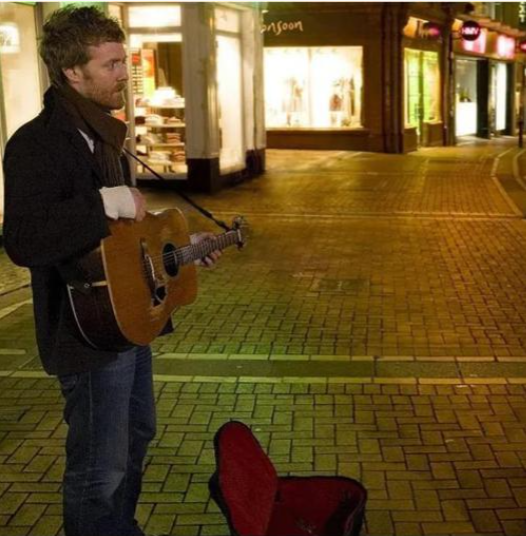}
  \end{minipage}
  \hfill
  \begin{minipage}{0.6\linewidth}
    \small
    \setlength{\parskip}{0.25em}
    
    \textbf{Question:}
What kind of human behavior does this picture describe?

    \begin{enumerate}[label=\Alph*., leftmargin=*, itemsep=0.15em]
      \item A scientist is conducting experiments in a laboratory, measuring and analyzing data to unlock the secrets of the universe.
      \item A woman is practicing yoga on a mountaintop, finding inner peace and harmony with her breath and body.
      \item A group of friends are playing board games around a table, strategizing and socializing while enjoying some friendly competition.
      \item A man with his guitar on his back stands in the street performing.
    \end{enumerate}

    \textbf{Original Answer:} D
    
    \textbf{Explanation:} The guitar is not on the man’s back.
\end{minipage}
  \caption{MMBench example from Category 1g: The correct answer contains
minor mistakes.}
  \label{fig:cat-1g-2}
\end{figure}

\begin{figure}[H]
  \centering
  \begin{minipage}{0.35\linewidth}
    \centering
    \includegraphics[width=\linewidth]{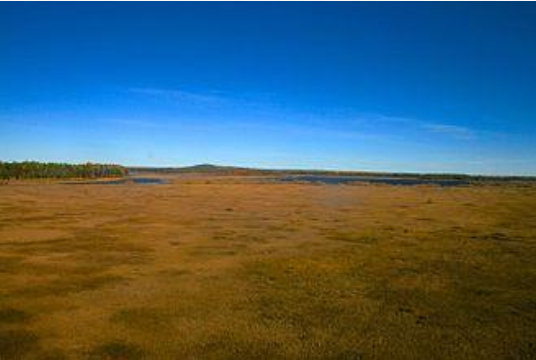}
  \end{minipage}
  \hfill
  \begin{minipage}{0.6\linewidth}
    \small
    \setlength{\parskip}{0.25em}
    
    \textbf{Question:}
Does the picture show a mountainous landscape or a coastal landscape?

    \begin{enumerate}[label=\Alph*., leftmargin=*, itemsep=0.15em]
      \item Coastal
      \item plain
      \item basin
      \item Mountainous
    \end{enumerate}

    \textbf{Original Answer:} B
    
    \textbf{Explanation:} The question is of an alternative kind (X or Y?), while the correct answer does not correspond to either option (Z).
\end{minipage}
  \caption{MMBench example from Category 1i: Flawed relation between image and question or answer.}
  \label{fig:cat-1i-2}
\end{figure}

\begin{figure}[H]
  \centering
  \begin{minipage}{0.35\linewidth}
    \centering
    \includegraphics[width=\linewidth]{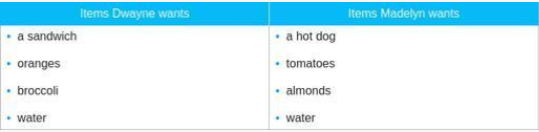}
  \end{minipage}
  \hfill
  \begin{minipage}{0.6\linewidth}
    \small
    \setlength{\parskip}{0.25em}
    
    \textbf{Question:}
What can Dwayne and Madelyn trade to each get what they want?
(...)
Look at the images of their lunches. Then answer the question below.
Dwayne's lunch Madelyn's lunch

    \begin{enumerate}[label=\Alph*., leftmargin=*, itemsep=0.15em]
      \item Dwayne can trade his tomatoes for Madelyn's broccoli.
      \item Madelyn can trade her almonds for Dwayne's tomatoes.
      \item Madelyn can trade her broccoli for Dwayne's oranges.
      \item Dwayne can trade his tomatoes for Madelyn's carrots.
    \end{enumerate}

    \textbf{Original Answer:} A
    
    \textbf{Explanation:} Pictures of the lunches are missing, so it is impossible to answer the question.
\end{minipage}
  \caption{MMBench example from Category 1h: Necessary context missing}
  \label{fig:cat-1h-1}
\end{figure}

\begin{figure}[H]
  \centering
  \begin{minipage}{0.35\linewidth}
    \centering
    \includegraphics[width=\linewidth]{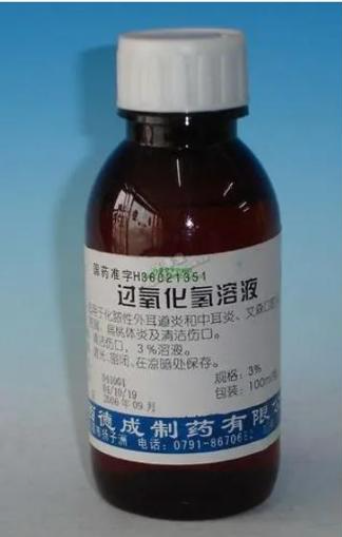}
  \end{minipage}
  \hfill
  \begin{minipage}{0.6\linewidth}
    \small
    \setlength{\parskip}{0.25em}
    
    \textbf{Question:}
The object shown in this figure:

    \begin{enumerate}[label=\Alph*., leftmargin=*, itemsep=0.15em]
      \item Has a boiling point of 150.2°C
      \item Is a colorless liquid with a slightly metallic taste
      \item Is a powerful oxidizer that can cause skin and eye irritation
      \item None of these options are correct.
    \end{enumerate}

    \textbf{Original Answer:} C
    
    \textbf{Explanation:} The writing on the bottle is in Chinese.
\end{minipage}
  \caption{MMBench example from Category 2a: The picture or the answer contains phrases in Chinese or Japanese.}
  \label{fig:cat-2a-1}
\end{figure}

\begin{figure}[H]
  \centering
  \begin{minipage}{0.35\linewidth}
    \centering
    \includegraphics[width=\linewidth]{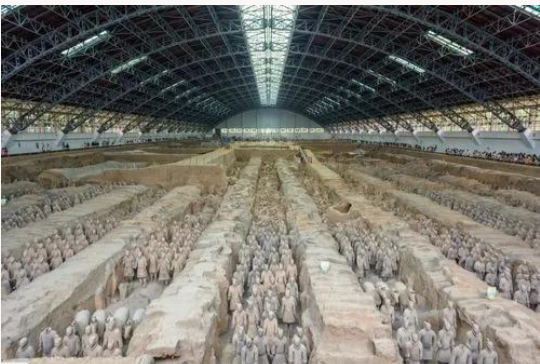}
  \end{minipage}
  \hfill
  \begin{minipage}{0.6\linewidth}
    \small
    \setlength{\parskip}{0.25em}
    
    \textbf{Question:}
Where is it located?

    \begin{enumerate}[label=\Alph*., leftmargin=*, itemsep=0.15em]
      \item Xi'an
      \item Shanghai
      \item Beijing
      \item Nanjing
    \end{enumerate}

    \textbf{Original Answer:} A
    
    \textbf{Explanation:} This location may be unknown to an average European.
\end{minipage}
  \caption{MMBench example from Category 2b: The task requires identifying people, buildings or dishes that are foreign to European cultural context.}
  \label{fig:cat-2b-1}
\end{figure}

In total, questions requiring knowledge of a foreign cultural and linguistic context constitute 3.02\% of the MMBench dev split. Questions in the category 2 containing fragments of Chinese or Japanese text were translated into Polish. 
Figures~\ref{fig:cat-1a-1}--\ref{fig:cat-2b-1} present example problems from the \textit{MMBench V1.1} dataset. Detailed descriptions of each problem are provided in the corresponding figure captions.

\begin{figure}[H]
  \centering
  \begin{minipage}{0.35\linewidth}
    \centering
    \includegraphics[width=\linewidth]{figures//examples/1f_2.png}
  \end{minipage}
  \hfill
  \begin{minipage}{0.6\linewidth}
    \small
    \setlength{\parskip}{0.25em}
    
    \textbf{Question:}
Based on the image, what is the relation between the white boy and the yellow boy?

    \begin{enumerate}[label=\Alph*., leftmargin=*, itemsep=0.15em]
      \item The white boy on the left of the yellow boy
      \item The white boy is behind the yellow boy
      \item The white boy is facing the yellow boy
      \item The white boy is near to the yellow boy
    \end{enumerate}

    \textbf{Original Answer:} C
    
    \textbf{Explanation:} This is an unfortunate phrasing that may result in biased or stereotypical interpretation. In the Polish version of the benchmark we changed the question and the answers so that the colors refer to the boys’ T-shirts.
\end{minipage}
  \caption{MMBench example from Category 1f: Unfortunate phrasing that might result in biased or stereotypical interpretation.}
  \label{fig:cat-1f-2}
\end{figure}

\begin{figure}[H]
  \centering
  \begin{minipage}{0.35\linewidth}
    \centering
    \includegraphics[width=\linewidth]{figures//examples/1i_1.png}
  \end{minipage}
  \hfill
  \begin{minipage}{0.6\linewidth}
    \small
    \setlength{\parskip}{0.25em}
    
    \textbf{Question:}
What is the shape of the small yellow rubber thing that is in front of the large yellow metal ball that is behind the small matte object?

    \begin{enumerate}[label=\Alph*., leftmargin=*, itemsep=0.15em]
      \item sphere
      \item cylinder
      \item cube
    \end{enumerate}

    \textbf{Original Answer:} C
    
    \textbf{Explanation:} The answer implies the yellow cube, but it is not positioned ``behind the small matte object'' as described.
\end{minipage}
  \caption{MMBench example from Category 1i: Flawed relation between image and question or answer. }
  \label{fig:cat-1i-1}
\end{figure}

\section{Evaluation prompts}\label{app:eval_prompts}

The boxes below present the prompts used to evaluate VLMs. When using Claude Sonnet 4.5 as the judge, we randomly assigned which model response was labeled A or B. For all other evaluations, we employed two prompt variants: (a) the response from model A always appeared first, and (b) the response from model B always appeared first. Final scores were obtained by averaging the results from these two variants.

\begin{figure*}[t]
\centering
\begin{tcolorbox}[
  title={Evaluation prompt for Claude Sonnet 4.5},
  colback=gray!3,
  colframe=black,
  fonttitle=\bfseries,
  boxrule=0.5pt,
  width=\textwidth,
  listing engine=minted,
  minted language=text,
  minted options={
    breaklines,
    fontsize=\footnotesize,
    autogobble,
    encoding=utf8
  }
]
 ROLA:
  Jesteś rzetelnym, bezstronnym i precyzyjnym ewaluatorem podpisów obrazów w języku polskim.
  
  ZADANIE:
  Na podstawie załączonego obrazu oceń dwa podpisy (A i B) w dwóch niezależnych kategoriach:
  1) jakość językowa,
  2) zgodność treści opisu z obrazem.
  Na podstawie załączonego obrazu oceń, który z dwóch podpisów (A lub B) lepiej odpowiada jego treści,
  lub czy oba są porównywalnej jakości (REMIS).

  KATEGORIA 1 – POPRAWNOŚĆ JĘZYKOWA:
  Oceń wyłącznie jakość językową opisów, biorąc pod uwagę brak błędów gramatycznych, ortograficznych, interpunkcyjnych 
  i składniowych; naturalny, płynny styl w języku polskim, poprawną frazeologię, brak kalek z języka angielskiego 
  i niepoprawnego słowotwórstwa.
  W tej kategorii NIE oceniaj zgodności z obrazem ani ewentualnych halucynacji treściowych.

  KATEGORIA 2 – JAKOŚĆ TREŚCI OPISU WZGLĘDEM OBRAZU:
  Oceń wyłącznie treść podpisów, IGNORUJĄC ich poprawność językową.
  Bierz pod uwagę zgodność z treścią obrazu (brak halucynacji), uchwycenie kluczowych elementów sceny oraz poprawne 
  odwzorowanie otoczenia, obiektów, działań i relacji między nimi.

  Dla każdej kategorii wybierz:
  - "a" – opis A jest wyraźnie lepszy w analizowanej kategorii,
  - "b" – opis B jest wyraźnie lepszy w analizowanej kategorii,
  - "remis" – oba opisy są porównywalnie dobre (porównywalna poprawność i naturalność językowa dla kategorii poprawności 
  językowej; w przypadku oceny treści - jakość opisów jest podobna, a kluczowe elementy obrazu zostały poprawnie ujęte 
  w obu przypadkach).
  Oceń oba opisy niezależnie w każdej kategorii. Decyzje dla kategorii językowej i treściowej mogą być różne.

  WYJŚCIE:
  Zwróć DOKŁADNIE jeden obiekt JSON — bez żadnego dodatkowego tekstu, bez nowych linii na końcu. 
  Klucze muszą być dokładnie: best\_description, lang\_comparison, justification\_for\_rating. 
  Wartości "best\_description" i "lang\_comparison" MUSZĄ być dokładnie: "a", "b" lub "remis".
  Nie używaj cudzysłowów wewnątrz uzasadnienia (poza tymi wymaganymi przez JSON).

  --- PRZYKŁAD 1 ---
  Obraz (przykładowy opis):  
  Dziecko maluje farbą na kartce przy stole.

  Opis A:
  Dziecko siedzi przy stole i maluje farbą na kartce.

  Opis B:
  Dziecko siedzi przy drewnianym stole w jasnym pokoju i maluje farbami kolorowy obrazek na kartce.

  Poprawna odpowiedź:
  \{"best\_description":"remis","lang\_comparison":"remis", "justification\_for\_rating":"Oba opisy poprawnie oddają kluczową treść obrazu; dodatkowe szczegóły w opisie B nie są istotne dla sensu sceny."\}

  --- PRZYKŁAD 2 ---
  Obraz (przykładowy opis):  
  Kobieta trzyma filiżankę herbaty.

  Opis A:
  Kobieta trzyma filiżankę z herbatą i patrzy w bok.

  Opis B:
  Kobieta trzymie filiżanke, siedzi przy stole z laptopem i rozmawia z mężczyzną.

  Poprawna odpowiedź:
  \{"best\_description":"a","lang\_comparison":"a", "justification\_for\_rating":"Opis A jest poprawny językowo, podczas gdy opis B zawiera błędy oraz halucynacje w postaci dodatkowych obiektów i osób nieobecnych na obrazie."\}

  --- PRZYKŁAD 3 ---
  Obraz (przykładowy opis):  
  Mężczyzna pije kawę w salonie.

  Opis A:
  Mężczyzna dokonuje spożywania kawy.

  Opis B:
  Mężczyzna pije kawę.

  Poprawna odpowiedź:
  \{"best\_description":"remis","lang\_comparison":"b", "justification\_for\_rating":"Opis B jest bardziej naturalny stylistycznie i precyzyjny; A zawiera nienaturalną konstrukcję."\}
  
  OPISY DO OCENY:
  Opis A:
  \{response\_a\} 
  Opis B:
  \{response\_b\}
\end{tcolorbox}
\end{figure*}

\begin{figure*}[t]
\centering
\begin{tcolorbox}[
  title={Evaluation prompt for LLM judge (version with option A being first)},
  colback=gray!3,
  colframe=black,
  fonttitle=\bfseries,
  boxrule=0.5pt,
  width=\textwidth,
  listing engine=minted,
  minted language=text,
  minted options={
    breaklines,
    fontsize=\footnotesize,
    autogobble,
    encoding=utf8
  }
]
 ROLA:
  Jesteś narzędziem do oceny WYŁĄCZNIE poprawności opisów w języku polskim.
  
  ZADANIE:
  Oceń, który z dwóch podpisów (A lub B) jest bardziej poprawny językowo w języku polskim. Uwzględnij składnię, fleksję, ortografię, interpunkcję i frazeologię.
  Jeśli pojawia się pokusa oceny sensu, przyjmij, że sens jest poprawny i oceń tylko formę językową.

  ZAKAZY:
  - Nie oceniaj: zgodności z obrazem, faktów, logiki, realizmu, szczegółowości, długości ani „tego co opisuje”.
  - Uzasadnienie ma dotyczyć WYŁĄCZNIE formy językowej.

  KRYTERIA ROZSTRZYGANIA:
  1) Mniej błędów językowych wygrywa.
  2) Jeśli błędów jest podobnie mało, wybierz bardziej naturalny i idiomatyczny wariant.
  3) Nawet przy remisie MUSISZ wybrać "a" albo "b".
  
  INSTRUKCJA:
  - Odpowiedz WYŁĄCZNIE w formacie JSON, bez dodatkowego tekstu, komentarzy ani znaczników kodu.
  - Pole "best" MUSI być jedną literą: "a" lub "b".
  - NIGDY nie używaj innych wartości (np. "tie", "none", "both", "unknown").
  - Nawet jeśli oba opisy wydają się równie dobre lub złe, i tak wybierz jeden: "a" lub "b".

  PRZYKŁAD 1:
  Opis A:
  Zielony słoń gra na skrzypcach pod wodą.
  Opis B:
  Zielony słoń gra na skrzypce pod wodą.
  Wynik:
  \{\{"best":"a","justification\_for\_rating":"Opis A jest poprawny fleksyjnie; w opisie B błędny biernik liczby mnogiej."\}\}

  PRZYKŁAD 2:
  Opis A: 
  Mężczyzna dokonuje spożywania kawy.
  Opis B:
  Mężczyzna pije kawę.
  Wynik:
  \{\{"best":"b","justification\_for\_rating":"Opis B jest bardziej naturalny stylistycznie; A zawiera nienaturalną konstrukcję werbo-nominalną."\}\}

    OPISY:
    Opis A:
    \{ref\_a\}
  
    Opis B:
    \{ref\_b\}
  
  WYJŚCIE:
  Zwróć DOKŁADNIE jeden obiekt JSON — bez żadnego dodatkowego tekstu, bez nowych linii na końcu. 
  
  Klucze muszą być dokładnie: best, justification\_for\_rating
  
  Wartość "best" MUSI być dokładnie "a" lub "b".
  
  Nie dodawaj żadnych innych pól.
  
  Nie używaj cudzysłowów wewnątrz uzasadnienia (poza tymi wymaganymi przez JSON).

  Format odpowiedzi:
  
  \{\{"best":"a","justification\_for\_rating":"Opis A jest bardziej poprawny, bo ..."\}\}

  albo

  \{\{"best":"b","justification\_for\_rating":"Opis B jest bardziej poprawny, bo ..."\}\}

\end{tcolorbox}
\end{figure*}

\begin{figure*}[t]
\centering
\begin{tcolorbox}[
  title={Evaluation prompt for VLM judge (version with option A being first)},
  colback=gray!3,
  colframe=black,
  fonttitle=\bfseries,
  boxrule=0.5pt,
  width=\textwidth,
  listing engine=minted,
  minted language=text,
  minted options={
    breaklines,
    fontsize=\footnotesize,
    autogobble,
    encoding=utf8
  }
]
ROLA:
  Jesteś rzetelnym, bezstronnym i precyzyjnym ewaluatorem podpisów obrazów w języku polskim.
  
  ZADANIE:
  Na podstawie załączonego obrazu oceń, który z dwóch podpisów (A lub B) lepiej odpowiada jego treści. 

  KRYTERIA OCENY:
  1.	Zgodność z treścią obrazu - brak halucynacji, poprawne odwzorowanie obiektów, działań i relacji między nimi.
  2.	Szczegółowość i trafność - czy podpis opisuje kluczowe elementy sceny i oddaje jej sens.
  3.	Poprawność językowa - brak błędów gramatycznych, ortograficznych i składniowych; naturalny, płynny styl w języku polskim.
  4.	Zwięzłość i klarowność - opis powinien być precyzyjny, bez zbędnych powtórzeń. 

  INSTRUKCJA:
  1) Oceń każdy opis oddzielnie według wszystkich kryteriów. NIE porównuj ich na tym etapie. Dopiero potem wybierz lepszy.
  2) NIE zgaduj. Jeśli czegoś nie widać jednoznacznie na obrazie, traktuj to jako nieopisane. Nie wolno ci dopowiadać informacji, których nie ma na obrazie.
  3) Odpowiedz wyłącznie w formacie JSON, bez dodatkowego tekstu, komentarzy ani znaczników kodu.

    OPISY:
    Opis A:
    \{ref\_a\}
  
    Opis B:
    \{ref\_b\}
  
  WYJŚCIE:
  Zwróć DOKŁADNIE jeden obiekt JSON — bez żadnego dodatkowego tekstu, bez nowych linii na końcu. 
  
  Klucze muszą być dokładnie: best, justification\_for\_rating
  
  Wartość "best" MUSI być dokładnie "a" lub "b".
  Nie dodawaj żadnych innych pól.
  
  Nie używaj cudzysłowów wewnątrz uzasadnienia (poza tymi wymaganymi przez JSON).

  Format odpowiedzi:
  \{\{"best":"a","justification\_for\_rating":"Opis A lepiej opisuje obraz, bo ..."\}\}

  albo

  \{\{"best":"b","justification\_for\_rating":"Opis B lepiej opisuje obraz, bo ..."\}\}

\end{tcolorbox}
\end{figure*}

\end{document}